\documentclass{article}
\usepackage{mystyle}
\newcommand{\parens}[1]{\left( #1 \right)}
\newcommand{\brackets}[1]{\left[ #1 \right]}
\newcommand{\braces}[1]{\left\{ #1 \right\}}

\newcommand{\set}[1]{\{ #1 \}}
\newcommand{\norm}[1]{\| #1 \|}

\newcommand{\R}{{\mathbb R}}

\DeclareMathOperator*{\E}{\mathbb{E}}
\newcommand{\expectation}[2]{\E_{#1}\brackets{#2}}
\newcommand{\kldivergence}[2]{\textbf{KL}\parens{#1 || #2}}

\newcommand{\Cat}[1]{\textbf{Cat}(#1)}
\newcommand{\Bern}[1]{\textbf{Bern}(#1)}

\newcommand{\Normal}[1]{\mathcal{N}(#1)}

\newcommand{\pftheta}{\Theta_{pf}}
\newcommand{\gmmtheta}{\Theta}

\newcommand{\elbo}{\text{ELBO}}

\newcommand{\GMM}{GMM}
\newcommand{\pfGMM}{pf-GMM}
\newcommand{\supGMM}{sup-GMM}
\newcommand{\pcGMM}{pc-GMM}
\newcommand{\twoGMM}{2-Step-GMM}

\newcommand{\HMM}{HMM}
\newcommand{\pfHMM}{pf-HMM}
\newcommand{\supHMM}{sup-HMM}

\newcommand{\twoHMM}{2-Step-HMM}

\newcommand{\logreg}{\text{LogReg}}

\newcommand{\zseq}{\textbf{Z}}
\newcommand{\xseq}{\textbf{X}}
\newcommand{\yseq}{\textbf{Y}}
\newcommand{\xiseq}{\boldsymbol{\xi}}
\newcommand{\seqn}[1]{{#1}^{(n)}}
\newcommand{\muvec}{\bm{\mu}}

\newcommand{\priorz}{{\theta}}
\newcommand{\paramtrans}{A}

\newcommand{\paramEmitRelevantMean}{B^{\mu}}
\newcommand{\paramEmitRelevantVar}{B^{V}}
\newcommand{\paramEmitRelevant}{B}
\newcommand{\paramsEmitRelevant}{{\paramEmitRelevant}}

\newcommand{\paramEmitIrrelevantMean}{\pi^{\mu}}
\newcommand{\paramEmitIrrelevantVar}{\pi^{V}}
\newcommand{\paramEmitIrrelevant}{\pi}
\newcommand{\paramsEmitIrrelevant}{{\paramEmitIrrelevant}}

\newcommand{\paramsY}{\eta^{\mu}, \eta^{V}}
\newcommand{\paramY}{\eta}

\newcommand{\data}{\xseq, \yseq}
\newcommand{\latentvars}{\zseq, \xiseq}
\newcommand{\miniposterior}{q(\latentvars)}
\newcommand{\fullposterior}{q(\latentvars | \Theta, \varphi)}

\newcommand{\drel}{$D_{\textbf{rel}}$}

\title{On Learning Prediction-Focused Mixtures}

\author[1]{Abhishek Sharma}
\author[1]{Catherine Zeng}
\author[1]{Sanjana Narayanan}
\author[1]{Sonali Parbhoo}
\author[1]{Finale Doshi-Velez}

\affil[1]{SEAS, Harvard University}
\date{}

\renewcommand{\emph}[1]{\textit{#1}}

\begin{document}

\maketitle

\begin{abstract}
  Probabilistic models help us encode latent structures that both model the data and are ideally also useful for specific downstream tasks.
  Among these, mixture models and their time-series counterparts, hidden Markov models, identify discrete components in the data.
  In this work, we focus on a constrained capacity setting, where we want to learn a model with relatively few components (e.g. for interpretability purposes).
  To maintain prediction performance, we introduce \emph{prediction-focused} modeling for mixtures, which automatically selects the dimensions relevant to the prediction task. Our approach identifies relevant signal from the input, outperforms models that are not prediction-focused, and is easy to optimize; we also characterize \emph{when} prediction-focused modeling can be expected to work.
\end{abstract}

\section{Introduction}
% 1. Context
% 2. Challenges
% 3. Our approach
Generative models provide a powerful mechanism for learning data distributions that can explain the data well and help predict specific outcomes of interest. For example, for patients with sepsis we might be interested in grouping clinical phenotypes and biological markers that are correlated with a clinical outcome such as mortality \parencite{seymouretal2019sepsis}.  In this context, a generative model elegantly handles missingness ---common in real-world datasets \parencite{cios2002uniqueness}--- and posits a mechanism for experts to understand where the model might be going wrong.

% FDV: I'm cutting this because I'm not sure if we need it in the narrative.  Either re-incorporate or delete.
% Despite these benefits, when it comes to predicting outcomes discriminative models often outperform generative models \parencite{ng2002discriminative}. %SP: I feel I'm missing the why here and how it ties to capacity

In this paper, we consider the task of learning a generative model that is also useful for predicting outcomes on some downstream tasks.  In particular, we focus on settings of \emph{limited capacity}, i.e. when the model does not have enough parameters to fully model the input distribution.  The limited capacity setting is relevant when we want models that are small enough to inspect, or in real-world settings where there will always be some level of misspecification.

Several methods have been proposed to learn generative models with good discriminative performance \parencite{Ganchev_Taskar_Gama_2008, blei_mcauliffe, ghahramani1994supervised}. The key challenge is to \emph{balance the trade-off} between the generative modeling objective and the prediction objective. The simplest two-step approach of maximizing the likelihood of the observed data and using that model for prediction largely fails because the learned model may not have modeled aspects of the inputs most useful for the prediction task. Alternatives such as prediction-constrained models \parencite{hughes2017predictionconstrained} explicitly model the generative-discriminative trade-off in a joint optimization, but are difficult to optimize in practice.

In our work, we manage the trade-off between explaining the data and the predictive objective by explicitly using `switch' variables that allow our models to select only those dimensions that are relevant to the downstream task, and enable accurate predictions even in settings where number of components is misspecified.
Unlike \cite{hughes2017predictionconstrained}, the entire problem can be written as a graphical model with the discriminative-generative trade-off being managed by a single hyperparameter corresponding to the probability that a dimension is relevant.  This setting was partially explored in \cite{ren2020prediction}, who considered a specific context of relevant words for topic models; we add the following contributions: 
% FDV: We can/should return to these one more time at the end before submitting, this is a pass to try to align what seems to be major contributions while also providing the context.  

\emph{We extend the idea of managing the generative-discriminative trade-off via relevant dimensions to continuous settings} (which do not have the orthogonality conditions used in \cite{ren2020prediction}): Gaussian Mixture Models (\GMM) and Hidden Markov Models (\HMM).  We demonstrate our approach not only yields performance benefits, but it is also more robust to optimize than more generic prediction constrained approaches.

\emph{We provide evidence that our model achieves the desired generative-discriminative trade-off directly from the maximum likelihood parameters of the associated graphical model}, instead of appealing to lower bounds or inference assumptions (as in \cite{ren2020prediction}).  In doing so, we also illuminate when this relatively simple, robust approach will work, and in what regimes it will fail.

% FDV: Commenting out for now.  Add back if we have space and there's something specific to say e.g. look to our analysis for XX, etc.
% We structure the paper as follows: in section \ref{sec:setting}, we introduce the problem setting; in sections \ref{sec:model} and \ref{sec:inference}, we introduce our model and a variational approximation to the posterior; in section \ref{sec:analysis}, we show using a synthetic example how our model's objective selects the input dimensions relevant for prediction; and finally in sections \ref{sec:experiments} and \ref{sec:results}, we demonstrate on synthetic and real datasets that prediction-focus trades off generative and predictive performance.

\section{Related Work}

\textbf{Semi-supervised Learning} aims to augment generative modeling of inputs \xseq\ with a supervisory signal $Y$ usually by training the joint objective $\log p(\xseq, Y)$ \parencite{NigamMTM98, kingma2014semisupervised, ghahramani1994supervised}. When modeling the generative process with latent variables \zseq\ as $\xseq \leftarrow \zseq \rightarrow Y$, this approach fails to recognize the inherent asymmetry between the inputs and the targets, and therefore ignores the target dimension as the model allocates its capacity to the more structured inputs. We address this issue by incorporating specific input-focused latent variables that treat the inputs and targets differently. Methods that incorporate an external signal to aid clustering of data exist under the paradigm of “supervised clustering” but they fail to provide a density of the observed data.

\textbf{Task-focused Generative Modeling} 
% FDV: Feel free to leave filling the cites til the end, but I think it'll be helpful to have... should be easy to pick some from the course :) 
Many recent works have considered ways to learn generative models that are useful for some downstream tasks \parencite{Lacoste–Julien_Huszár_Ghahramani_2011, Cobb_Roberts_Gal_2018, Stoyanov_Ropson_Eisner_2011, popcorn}. Most of these train for the discriminative loss only.  Closer to our goal of managing the generative-discriminative trade-off, \textcite{hughes2017predictionconstrained} proposed prediction-constrained learning in the context of mixture and topic models, noting the asymmetry of the task in predicting labels from documents (and not the other way around).  Their framework balances the maximum likelihood objective with predictive performance by introducing the latter as a constraint and then solving the resulting constrained optimization problem. But the resulting constrained objective is often hard to optimize in practice, and unlike our work, it does not correspond to maximum likelihood of a valid graphical model. 

In the context of topic models for text documents, \textcite{ren2020prediction} propose using switch variables to exclude or include words depending on their relevance for prediction.  Their approach relies on independence assumptions between switches and topic distributions in the approximate posterior for the model to learn useful topics.
% But they rely on orthogonality of relevant and irrelevant topic vectors for the model to work---a condition which is specific to topic models and discrete distributions of the words.
Our models do not rely on such conditions during inference, and have the prediction-focused properties by virtue of their structure alone.

% \paragraph{Information bottleneck methods}
\textbf{Information Bottleneck (IB) methods}
take a joint distribution $p(\textbf{X}, Y)$ and compress the inputs $\xseq$ into a latent code $\textbf{Z}$ that is most informative about the target $Y$ \parencite{tishby2000information}.
Both the traditional IB and the deep variational IB formulations assume specific Markov structures for the generative model \parencite{alemi2016deep, Wieczorek_2020}. In contrast, we achieve the trade-off between compressing inputs and predicting targets by proposing a specific probabilistic model which encodes our knowledge about the generative process. We demonstrate that this model is sufficient for getting the trade-off we seek, independent of the IB objective.

\section{Problem Setting}
\label{sec:setting}
We study the case where the inputs \xseq\ are a concatenation of $M$ independent mixtures, each with $K_m$ components ($m \in [M]$). Without loss of generality, we assume that the first mixture also generates our target of interest, $Y$:
\begin{gather*}
    \xseq = [\xseq_1; \dots; \xseq_M] \\
    [\xseq_1; Y] \sim \textsc{GMM}(K_1, \Theta_1); \quad \xseq_m \sim \textsc{GMM}(K_m, \Theta_m)
\end{gather*}
This is a setting which is pervasive in real world datasets, where there is often much more structure in the input data \xseq\ than is needed to make predictions for a specific task (e.g. predicting $Y$).  To perfectly model the input distribution $p(\xseq)$, we would need a very large generative model (with $K = \prod_m K_m$ components).  Such a large model might be hard to fit with limited data, and may be unnecessary if we are most interested in capturing structure in the data that is relevant to the task.  

For example, in a healthcare setting, different dimensions may correspond to sets of measurements related to different conditions such as kidney disease, mental health, or a broken leg.  If we are most interested in predicting outcomes related to recovery from a broken limb, then modeling variation in broken leg dimensions might be essential, whereas modeling variation in kidney disease may be less important (even if a chronic disease generates more data).

Below, we consider misspecification in the number of mixture components and ensure that the model prioritizes the use of components to learn those that are relevant. 
Formally, our setting pertains to the case where we are provided with a fixed budget K ( $< \prod_m K_m$) of components we can use, and we are required to (a) perform well on the task of predicting $Y$ at test time, and (b) return a generative model of data that achieves as high a value of $p(\xseq)$ as possible given that we also want to perform well on predicting $Y$.
% \vspace{-0.4cm}
\section{Prediction-Focused Mixtures}
\label{sec:model}
\begin{figure*}[t]
    \centering
    \begin{subfigure}[b]{0.4\textwidth}
        \begin{tikzpicture}
        % \buildnodes
        %% NODES
        %% z
        \node[latent] (zn) {$Z^{(n)}$} ; %
        \node[const, left=of zn] (priorz) {$\priorz$} ; %
        %% x
        \node[obs, above=of zn] (xnd) {$X_d^{(n)}$} ; %
        \node[const, right=2.5cm of xnd] (paramsX) {$B, \pi$} ; %

        %% y
        \node[obs, right=of zn] (yn) {$Y^{(n)}$} ; %
        \node[const, right=of yn] (paramsY) {$\paramY$} ; %

        %% xi
        \node[latent, left=of xnd]  (xid) {$\xi_{d}$};

        %% p
        \node[const, below=of xid] (p) {$p$} ; %

        \plate {platedims} {
            (xnd)
            (xid)
        } {$D$}
        
        \plate {} {
            % (platedims)
            (xnd)
            (zn)
            (yn)
        } {$N$}

        %% Edges

        %% z_n -> x_nd
        \edge {zn} {xnd} ; %
        % %% xi_d -> x_nd
        \edge {xid} {xnd} ; %
        %% paramsX -> x_td
        \edge {paramsX} {xnd} ; %

        % %% z_n -> y_n
        \edge {zn} {yn} ; %
        \edge {paramsY} {yn} ; %

        %% p -> xi_d
        \edge {p} {xid} ;

        % %% theta -> z_n
        \edge {priorz} {zn} ;
        \end{tikzpicture}%
    \caption{} \label{fig:pfGMM}
    \end{subfigure}
    \begin{subfigure}[b]{0.5\textwidth}
        \begin{tikzpicture}
        % \buildnodes

        %% NODES
        %% z
        \node[latent] (z1) {$Z_1$} ; %
        \node[latent, right=of z1] (z2) {$Z_2$} ; %
        \node[const, right=of z2] (zdots1) {$\dots$} ; %
        \node[latent, right=of zdots1] (zt) {$Z_t$} ; %
        \node[const, right=of zt] (zdots2) {$\dots$} ; %
        \node[latent, right=of zdots2] (zT) {$Z_T$} ; %
        \node[const, below=of z2] (paramtrans) {$\paramtrans$} ; %
        \node[const, below=of z1] (priorz) {$\priorz$} ; %
        
        %% xi
        \node[latent, above=of z2]  (xid) {$\xi_{d}$};
        
        %% x
        \node[obs, right=of xid] (xtd) {$X_{t, d}$} ; %
        \node[const, right=of xtd] (paramsX) {$B, \pi$} ; %

        %% y
        \node[obs, right=of paramsX] (yt) {$Y_t$} ; %
        \node[const, right=of yt] (paramsY) {$\paramY$} ; %

        %% p
        \node[const, left=of xid] (p) {$p$} ; %

        \plate {name} {
            (xtd)
            (xid)
        } {$D$}
        
        %% Edges

        %% z_i -> x_td
        \edge {zt} {xtd} ; %
        %% xi_td -> x_td
        \edge {xid} {xtd} ; %
        %% paramsX -> x_td
        \edge {paramsX} {xtd} ; %

        %% z_i -> y_i
        \edge {zt} {yt} ; %
        \edge {paramsY} {yt} ; %
        
        %% p -> xi_td
        \edge {p} {xid} ;

        \edge {z1} {z2} ;
        \edge {z2} {zdots1} ;
        \edge {zdots1} {zt} ;
        \edge {zt} {zdots2} ;
        \edge {zdots2} {zT} ;
        \edge {priorz} {z1} ;
        \edge {paramtrans} {z2} ;
        \edge {paramtrans} {zt} ;
        \edge {paramtrans} {zT} ;
        
        \end{tikzpicture}%
    \caption{} \label{fig:pfHMM}
    \end{subfigure}
    \caption{The (a) \pfGMM\ and (b) \pfHMM\ graphical models}
\end{figure*}
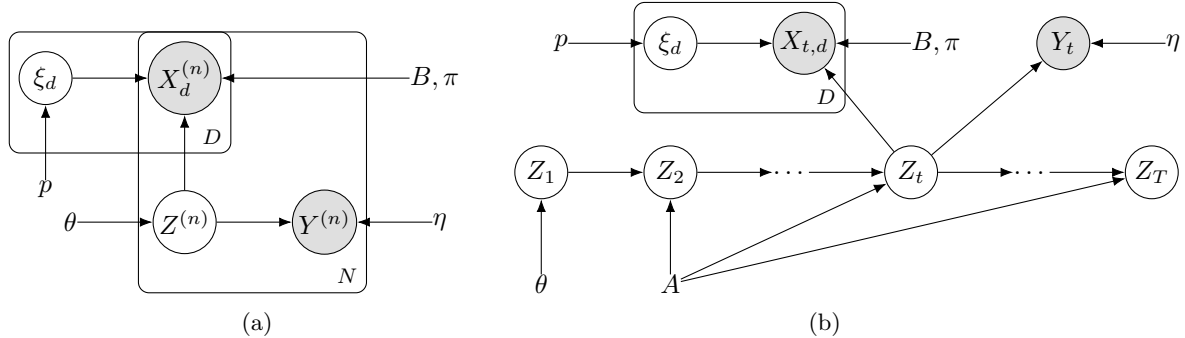
We now present prediction-focused modeling, first by introducing prediction-focused Gaussian Mixture Models (\pfGMM) (Fig.~\ref{fig:pfGMM}) and then extending it to time series setting by introducing prediction-focused Hidden Mixture Models (\pfHMM) (Fig.~\ref{fig:pfHMM}).

The generative process of the model is:

\begin{algorithmic}
    \FOR{each dimension $d$}
    \STATE Draw switch $\xi_{d} \sim \Bern{p}$
    \ENDFOR
    \FOR{each datum $n$}
        \STATE Draw latent state $Z^{(n)} \sim \Cat{\theta}$
            \FOR{each dimension $d$}
            \IF{$\xi_{d} = 1$}
                \STATE Draw `signal' obs.\\ $X_d^{(n)} \sim F_X(X_d^{(n)} | Z^{(n)}; \paramEmitRelevant_{Z^{(n)},d})$
            \ELSIF{{$\xi_{d} = 0$}}
                \STATE Draw `noise' obs. $X_d^{(n)} \sim F_X(X_d^{(n)}; \paramEmitIrrelevant_{d})$
            \ENDIF
        \ENDFOR
        \STATE Draw target $Y_d^{(n)} \sim F_Y(Y_d^{(n)};\paramY_{Z^{(n)}})$
    \ENDFOR
\end{algorithmic}
where $\Bern{p}$ is a Bernoulli distribution with parameter $p$, $\Cat{\theta}$ is Categorical distribution with parameter vector $\theta$, and $F_X, F_Y$ can be any distribution over the support of $X_d^{(n)}$. In our paper, we choose the inputs to be Gaussian for simplicity of exposition, and hence refer to the model as \pfGMM.
\pfGMM\ filters out `irrelevant' dimensions by the use of switch variables $\set{\xi_d}$, allowing it to use only a subset of dimensions to model along with the targets. This property allows it to achieve good downstream performance even when it is (inevitably) misspecified w.r.t. the data generating process. Here, we assume that the input data dimensions can be partitioned as being either `relevant' or `irrelevant', where `relevant' dimensions are predictive of the target, $Y$.

% \paragraph{Hyperparameters}
% The \pfGMM\ requires us to tune only a single hyperparameter: the switch prior $p$.
% Intuitively, it encodes our belief about roughly how many dimensions should be relevant. But from a model training point of view, it also trades off generative quality with predictive performance.

% \paragraph{Robustness to misspecification}
% Any real world analysis faces a situation where the proposed model's generative process doesn't match the true data generating process. We design the \pfGMM\ to be robust to some of such misspecifications. While an arbitrarily high number of components can explain the data well, it is reasonable to expect only a few to be useful for the prediction task.
% We consider misspecification in the number of mixture components and ensure that the model prioritizes the use of components to learn those that are relevant. This property seeks out mixture components that are relevant to the task at hand, instead of just picking out those that have more structure. 
% \note{redundant after the section \ref{sec:setting}?}

\section{Inference}
\label{sec:inference}

Since \pfGMM\ is a valid graphical model, we can use the maximum likelihood objective to do learning and inference. The log likelihood of the \pfGMM\ is: 
\begin{align}
    \log p  (\data) &= \log \sum_{\latentvars} p_{\theta}(\zseq) p_{p}(\xiseq) p_{B,\pi}(\xseq | \latentvars) p_{\eta}(\yseq | \zseq)
    \label{eq:ll}
\end{align}
Computing Eq.~\ref{eq:ll} is generally intractable, but in Sec.~\ref{sec:analysis}, we demonstrate that in lower dimensional settings (where the computation is tractable), maximizing this likelihood gives good discriminative performance---suggesting that the \pfGMM\ model \emph{itself} balances the generative-discriminative trade-off.

In practice, we optimize a lower bound (the \elbo) to Eq.~\ref{eq:ll}. We choose the following form for the approximate posterior:
\begin{align*}
    \fullposterior &= \prod_{n=1}^N q(Z^{(n)} | \xseq^{(n)}, Y^{(n)}, \varphi, \Theta) \prod_{d=1}^D q(\xi_d^{(n)} | \varphi_d)
\end{align*}
where $\Theta = \set{\priorz, \paramEmitRelevant, \paramEmitIrrelevant, \paramY}$ refers to the model parameters.  In particular, note that we have added an additional index to the switch variable; now $\xi_d^{(n)}$ is a function of the datum $n$ as well as the dimension $d$.  This will allow for a cleaner factorization; we model the posterior Bernoulli distributions $q(\xi_d^{(n)} | \varphi_d)$ with a \emph{tied} parameter $\varphi_d$ for dimension $d$ to still get a per-dimension relevance.

The distribution $q(Z^{(n)} | \xseq^{(n)},Y^{(n)}, \varphi, \Theta)$ is simply the posterior distribution conditioned on the variational parameters $\varphi$ and can be computed as:
\begin{align*}
    q(Z = k | \xseq,Y, \varphi) &\propto 
    p_{\priorz}(Z = k) \cdot p(\xseq | Z = k, \varphi) \cdot p_{\paramY}(Y | Z = k)
\end{align*}
At prediction time, the distribution $q(Z = k | \xseq,Y, \varphi)$ is computed without the $p_{\paramY}(Y | Z = k)$ term and the \textit{feature selection parameter} $\varphi$ helps select the right features in the term $p(\xseq | Z = k, \varphi)$.

This results in the following \elbo\ objective (full derivation in Supplement Sec. 1): 
\begin{align}
    \elbo(q) &= \expectation{\miniposterior}{\log p_{\Theta}(\data | \latentvars)} \notag\\
    &\quad\quad\quad - \kldivergence{\fullposterior}{p_{\Theta}(\zseq, \xiseq)}
    \label{eq:elbo}
\end{align}
which is tractable to compute because the posterior factorizes over $\xiseq$ and $\zseq$. After this setup, the model can be trained using the variational EM algorithm which alternates between learning the parameters $\Theta$ and inferring the approximate posterior $\fullposterior$.

Similarly, we train \pfHMM\ using variational inference by using the following approximate posterior, $\fullposterior$:
\begin{align*}
&= \prod_{n=1}^N \parens{\prod_d \prod_t q(\seqn{\xiseq}_{t,d} | \varphi_d)} p(\seqn{\zseq} | \seqn{\xseq},\seqn{\yseq}, \varphi, \Theta)
\end{align*}
where the subscript $t$ indexes time within a sequence and $\seqn{\zseq}, \seqn{\xseq},\seqn{\yseq}$ are variables corresponding to sequence $n$.
Training proceeds using variational EM, with the only exception that we now use the Forward-Backward algorithm to infer $p(\seqn{\zseq} | \seqn{\xseq},\seqn{\yseq}, \varphi, \Theta)$ during the E-step. We provide the complete derivation for the \elbo\ in Supplement Sec. 1.

\section{Analysis}
\label{sec:analysis}

\begin{figure*}[t]
    \centering
    \begin{subfigure}[t]{0.3\textwidth}
        \includegraphics[width=\textwidth]{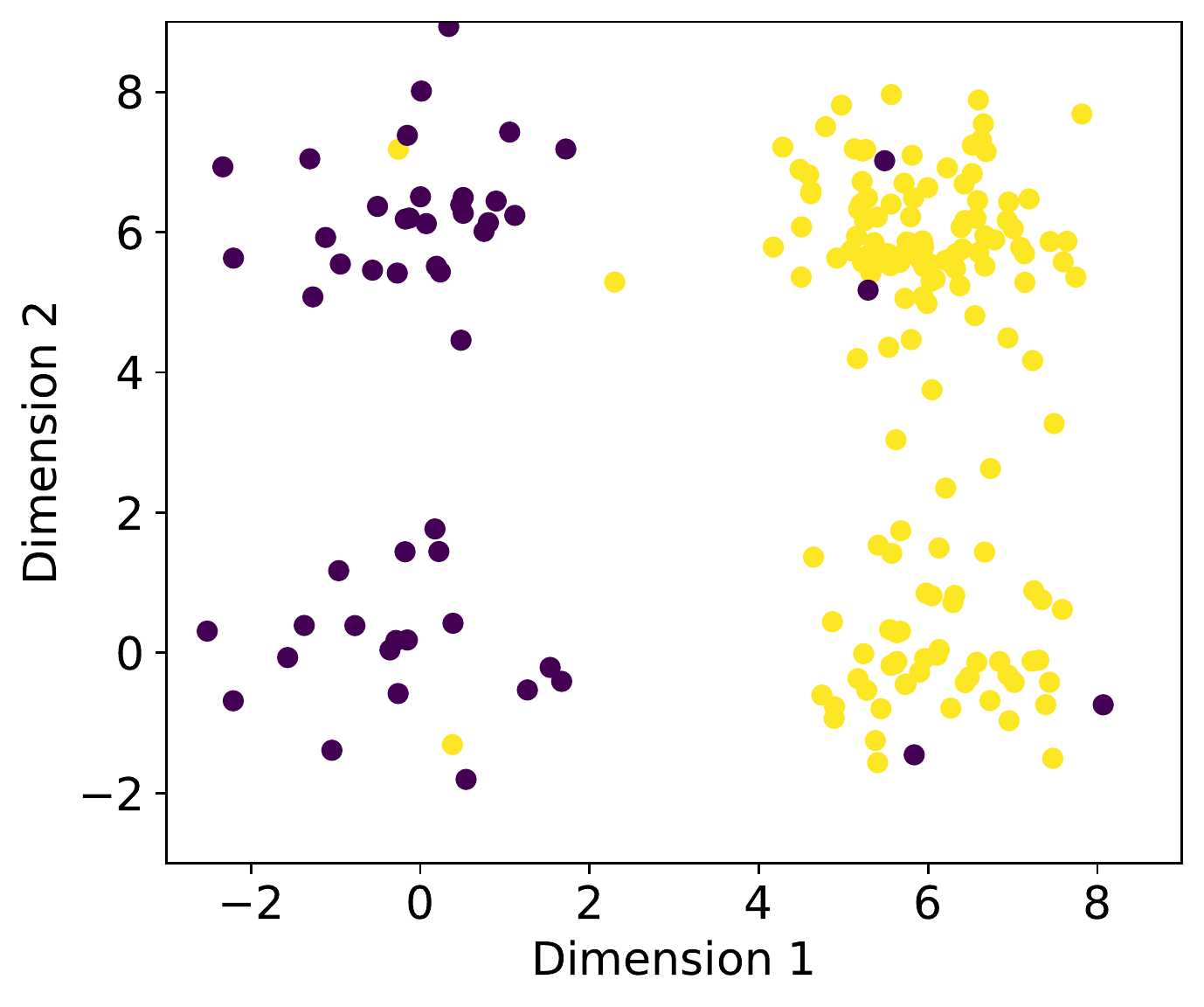}
        \caption{}
        \label{fig:analysis0}
    \end{subfigure}
    \begin{subfigure}[t]{0.3\textwidth}
        \includegraphics[width=\textwidth]{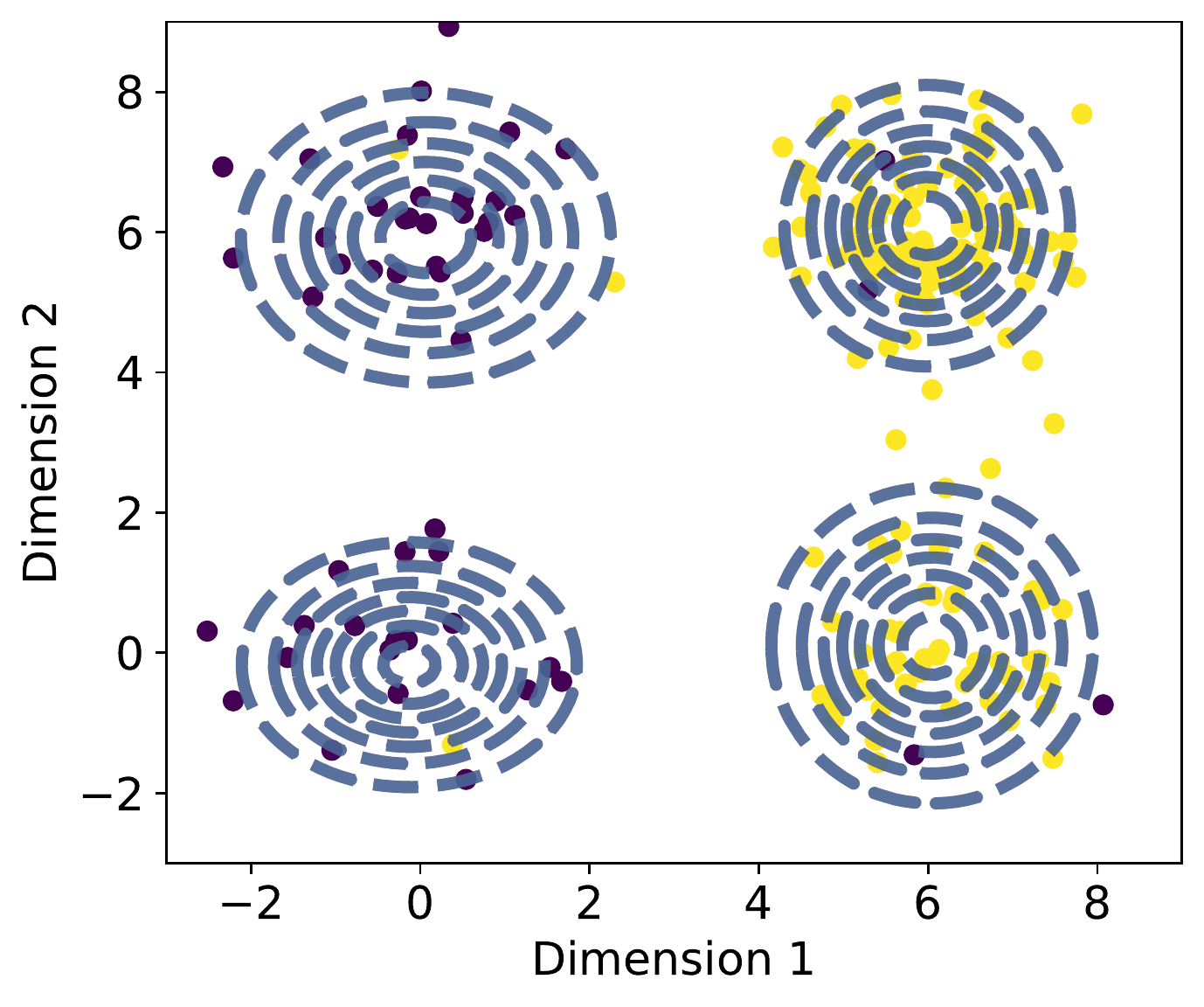}
        \caption{}
        \label{fig:analysis1}
    \end{subfigure}
    \begin{subfigure}[t]{0.3\textwidth}
        \includegraphics[width=\textwidth]{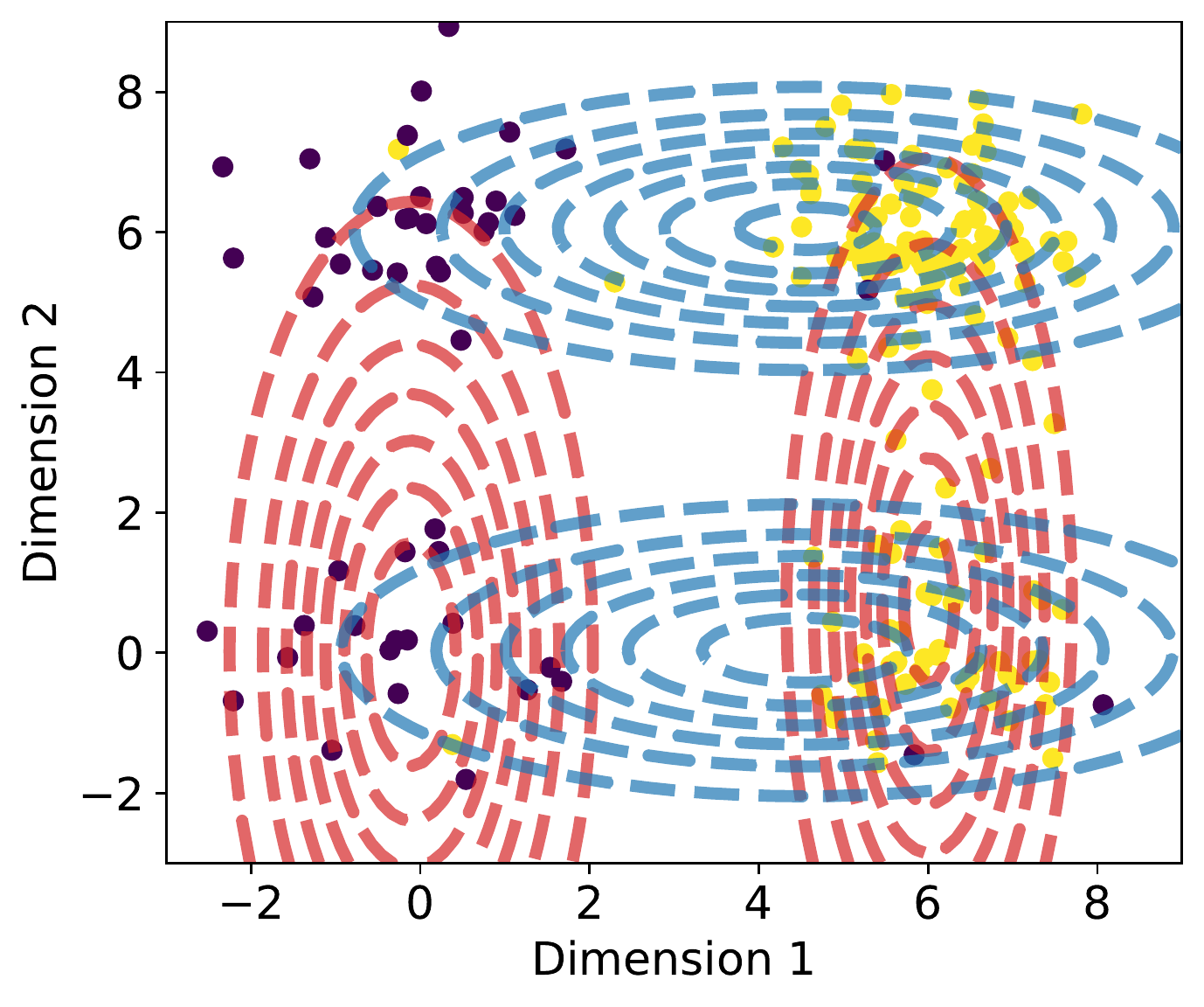}
        \caption{}
        \label{fig:analysis2}
    \end{subfigure}
    \begin{subfigure}[t]{0.3\textwidth}
        \includegraphics[width=\textwidth]{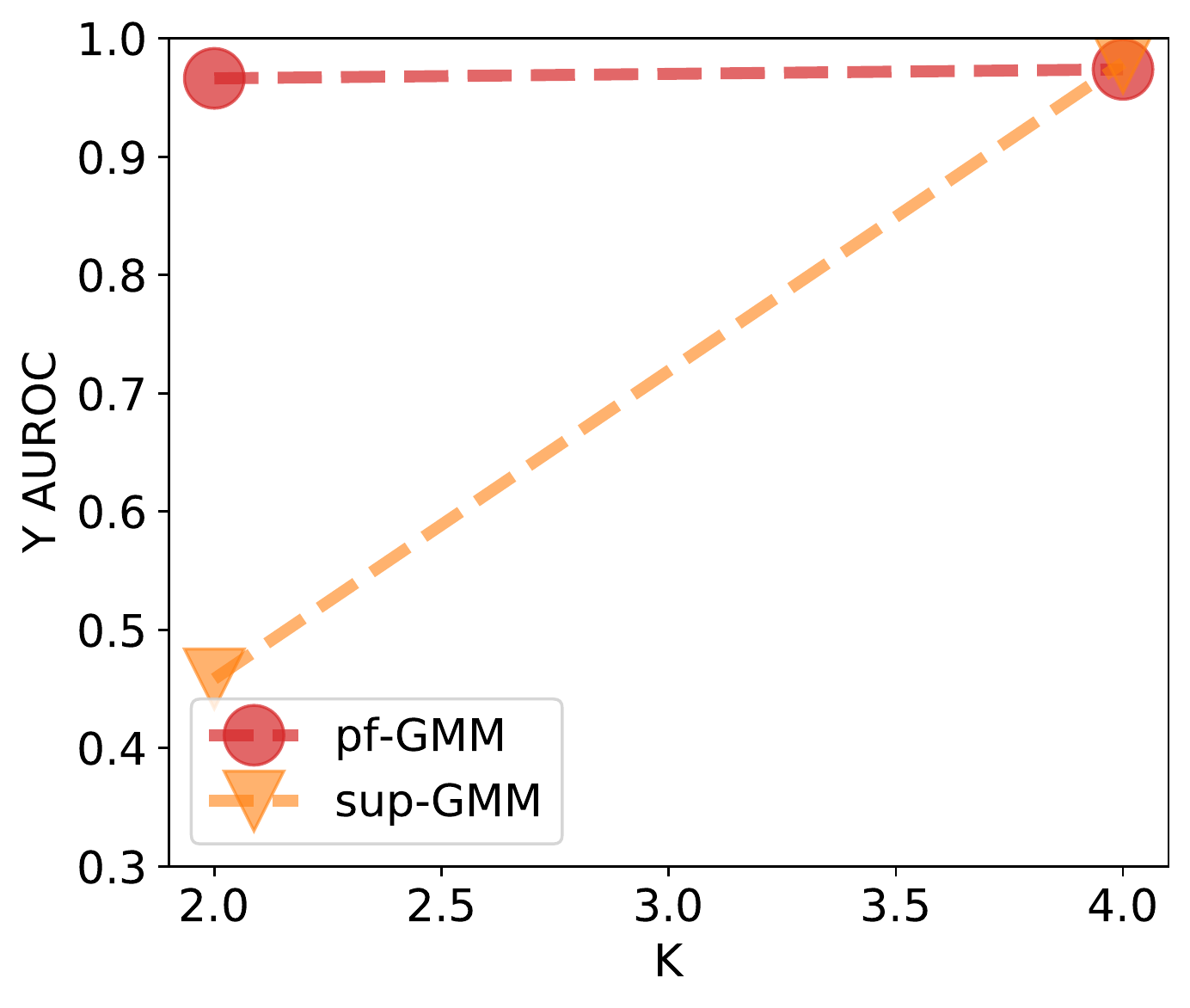}
        \caption{}
        \label{fig:analysis3}
    \end{subfigure}
    \begin{subfigure}[t]{0.3\textwidth}
        \includegraphics[width=\textwidth]{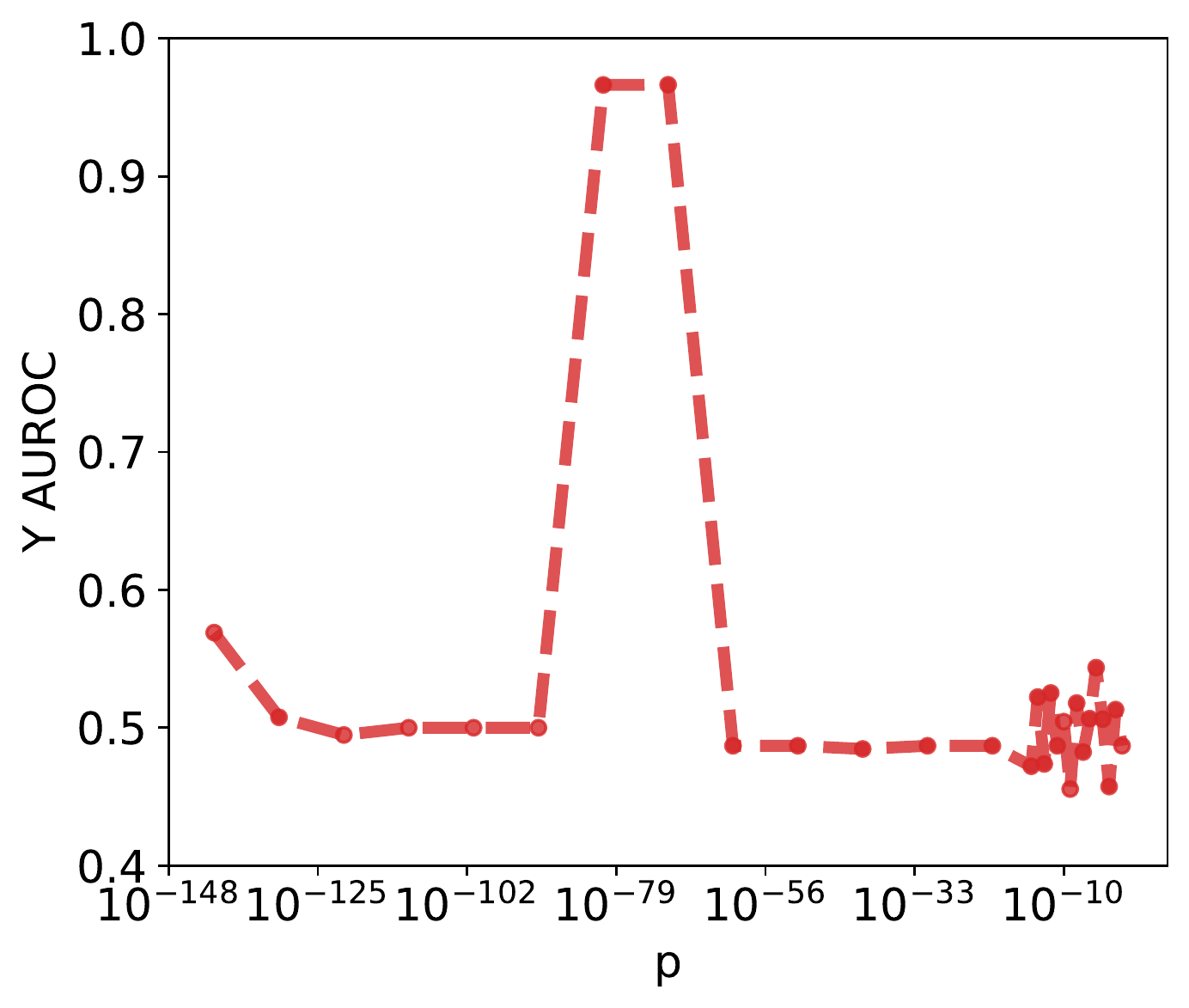}
        \caption{}
        \label{fig:analysis4}
    \end{subfigure}
    \begin{subfigure}[t]{0.3\textwidth}
        \includegraphics[width=\textwidth]{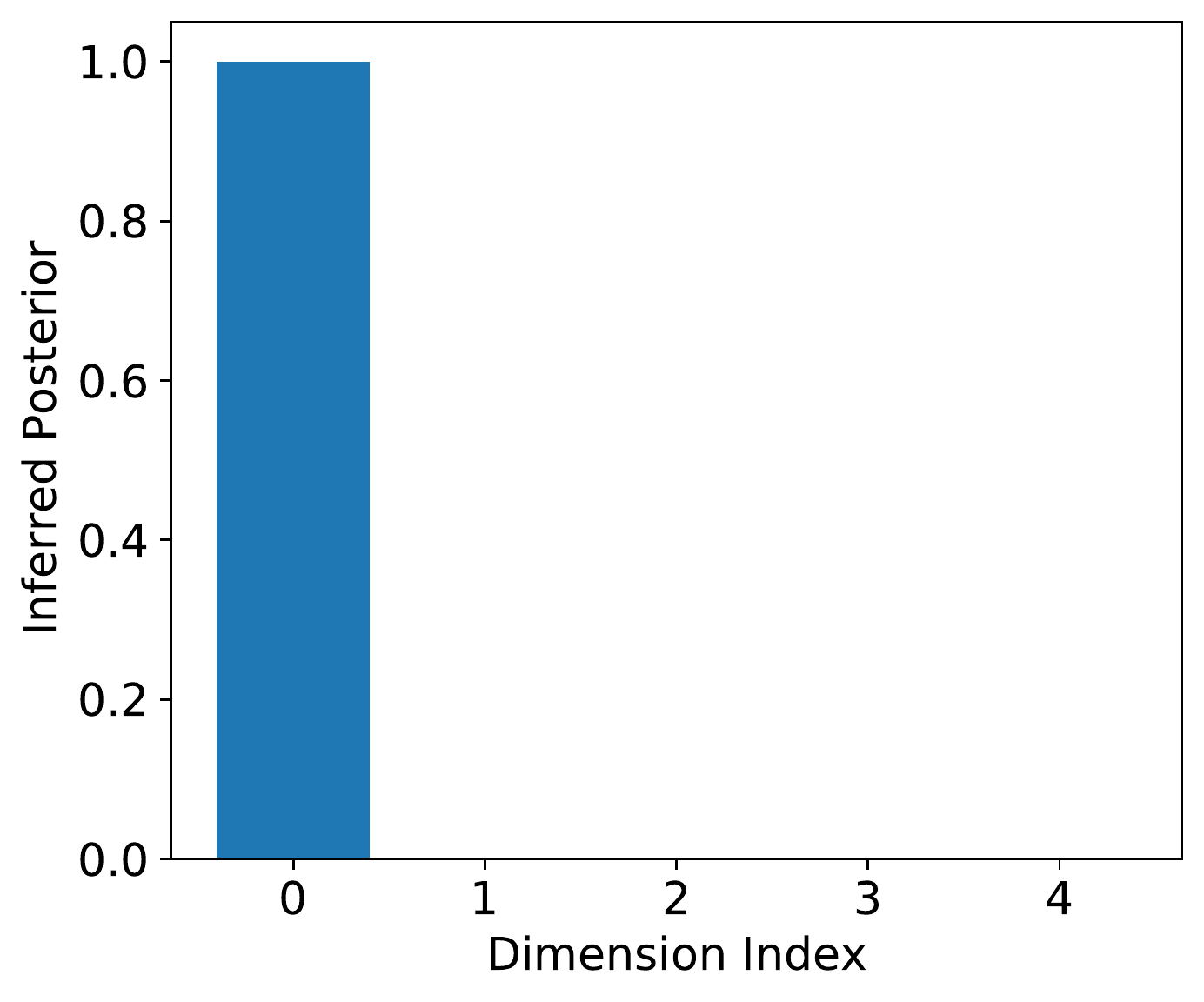}
        \caption{}
        \label{fig:analysis5}
    \end{subfigure}
    \caption{Comparison of \pfGMM\ with a supervised \GMM\ (\supGMM) on a synthetic dataset. (a) First 2 (out of 6) dimensions of the input data, with colors identifying the target label. (b) Learned components of both models when the component budget is 4 (identical in this case). (c) Learned components of both models restricted to 2 components---\pfGMM\ (red contours) learns components that are predictive of the labels whereas \supGMM\ (blue) components achieve higher $p(\xseq)$ because they align with more input dimensions. (d) Heldout $Y$ AUROC (predictive performance) vs component budget (K)---\pfGMM\ performs better under misspecification and no worse than \supGMM\ otherwise. (e) Heldout AUROC vs \pfGMM\ hyperparameter $p$---performance peaks at an optimal $p$ and drops when it is higher (more dimensions considered as signal) or lower (no dimension considered as signal). (f) Inferred switch posterior for the case when component budget is 2 (4-component case is the same)---\pfGMM\ identifies the relevant dimension.}
\end{figure*}

Earlier work motivated using a switch-variable based approach to relevant and irrelevant dimensions by noting that Eq.~\ref{eq:ll} can also be lower bounded by the following \textcite{ren2020prediction} (Supplement 9.4): 
\begin{align}
    \log p(\data) &\geq \expectation{p(\xiseq)}{\log p(\yseq | \xseq, \xiseq)} + p \expectation{p(\zseq)}{\log p_{\paramEmitRelevant}(\xseq | \zseq)} \notag \\
    & \quad + (1-p) \log p_{\paramEmitIrrelevant}(\xseq)
    \label{eq:altlb}
\end{align}
While this bound does explicitly demonstrate a generative-discriminative trade-off analogous to \textcite{hughes2017predictionconstrained}, the bound can neither be optimized directly in a computationally efficient manner, nor does it imply that the joint likelihood in Eq.~\ref{eq:ll} captures this trade-off---indeed, it would be unfortunate if our results were because of our choice of approximate inference rather than the due to the model itself.

% FDV: I'm confused by the below: Shouldn't the limitation be mentioned here?  I feel like that's important, because that's a difference between the claim in Jason's work and ours (it will work in many regimes, but will fail when... this is valuable because it provides insight into when using this kind of trick to simplify the generally complex task-focused learning problem will be helpful...) 
In this section, we demonstrate using a synthetic example that the maximum likelihood estimate (MLE) of the actual \pfGMM\ performs at least as well as the \GMM\ at the prediction task. We also characterize a limitation of prediction-focused modeling that provides insight into the data regimes in which we would expect this style of modeling to work. Specifically we show that (a) the \pfGMM\ solution is equivalent to the \GMM\ solution when there is no misspecification, (b) the \pfGMM\ MLE is better than the \GMM\ MLE solution when there is misspecification in the number of clusters, and (c) \pfGMM\ MLE is no better than than the \GMM\ MLE when noise dimensions cannot be modeled using a simple distribution, $F_X(\cdot ; \paramEmitIrrelevant_{d})$.

We choose an example that is instructive of the setting where \pfGMM\ would do well: (a) the input data has structured noise coming from a mixture of Gaussians, and (b) only a small proportion of the data dimensions is `signal'. This simulates a setting in which `noise' competes with the `signal' for parameters, and na\"ive joint modeling of inputs and targets is not expected to do well. 
We assume the data comes from the following generative process:
\begin{gather*}
    [X_1, y] \sim 0.5 [\Normal{0, 1}, \Bern{0}] + 0.5 [\Normal{0, 1}, \Bern{1}] \\
    \xseq_2 \sim 0.5 \Normal{\textbf{0}, I} + 0.5 \Normal{\textbf{6}, I} ; \quad \xseq_2 \in \R^4 \\
    \xseq = [\xseq_1, \xseq_2]
\end{gather*}
where \textbf{0} and \textbf{6} are constant vectors in $\R^4$. In this data, only the first dimension of $X$ is correlated with output $y$ while the rest of the dimensions are not. Fig. \ref{fig:analysis0} plots the first two dimensions of independent samples from this distribution. We clearly see that we need four clusters to model the data well and also predict targets correctly.
Next, we see the nature of solutions that are achieved using \pfGMM\ and \GMM\ for different component budgets.
Finally, we will characterize a data setting where we would not expect \pfGMM\ model to do better than a \GMM.

\paragraph{No misspecification}
Clearly, 4 components are sufficient to model the inputs accurately in this problem. Therefore, given a budget of 4 components, the GMM will identify each of the emission distributions, and at test time, will predict the output labels correctly. Also note that the \pfGMM\ with $p=1$ will be equivalent to the GMM, which is what we see when we look at the learned components in Fig. \ref{fig:analysis1}.

\paragraph{Misspecification}
If we are restricted to a budget of only 2 components, the GMM identifies clusters that align with the majority of input dimensions rather than aligning itself to the first dimension (Fig. \ref{fig:analysis2}). This is true because it is trying to do well on the joint likelihood, $p(\xseq,y) = p(\xseq)p(Y|\xseq)$. In the case of GMM, the benefit from modeling $p(\xseq)$ well is much higher than cost of modeling $p(Y|\xseq)$ poorly. The \pfGMM\ bypasses this issue by tuning $p$ to a value that selects only a subset of dimensions to be relevant and explains the rest using its noise distribution, $F_X(X_d; \paramEmitIrrelevant_{d})$. We can therefore see in Fig. \ref{fig:analysis4}, that for an appropriate value of $p$, it selects the clusters that capture the variation along outputs, leading to high predictive accuracy.

\paragraph{Limitation of \pfGMM}
The hyperparameter $p$ is useful in tuning the number of effective dimensions that \pfGMM\ chooses to model using its components, trading off generative performance with discriminative performance. Nonetheless, this causes \pfGMM s to have a limitation which we characterize next. Suppose the irrelevant dimensions were instead generated using the following distribution:
\begin{align*}
    \xseq_2 &\sim 0.5 \Normal{\textbf{0}, I} + 0.5 \Normal{\muvec, I} ; \quad \xseq_2 \in \mathbb{R}^4
\end{align*}
where \textbf{0} and $\muvec$ are constant vectors in $\R^4$, and $\mu$ is the distance between the component means of each of the irrelevant dimensions (previously, $\muvec$ was \textbf{6}).
Now consider the case when $\mu$ becomes large, i.e. the noise components provide a much higher contribution to $p(\xseq)$ for any dimension. Let the parameters learned with and without prediction-focus in the above example be $\pftheta$ and $\gmmtheta$ respectively. Then as $\mu \rightarrow \infty$, their likelihood ratio goes to 0, making it asymptotically beneficial to select $\gmmtheta$ as the maximum likelihood solution. We prove this claim in Supplement Sec. 2, but the intuition is that the \pfGMM\ selects the number of relevant dimensions. However, which dimensions those are will depend on how much they contribute to the generative term, $p(\xseq)$. If any dimension takes a \textit{much} bigger hit when modeled as noise (vs as signal) than other dimensions, \pfGMM\ will find it hard to classify it as noise.

% This analysis highlights the assumptions we make about the data for \pfGMM\ to give better predictive performance. 
This analysis highlights in which ways a \pfGMM\ can be misspecified and still give good predictive performance, and in which ways it cannot; specifically, when each dimension can be reasonably modeled using a fixed Gaussian (i.e. $\Normal{X_d; \mathbb{E}[X_d], \mathbb{V}[X_d]}$ does not go to zero), \pfGMM\ will do well.
% Second, we assume that each dimension can be reasonably modeled using a fixed Gaussian (i.e. $\Normal{X_d; \mathbb{E}[X_d], \mathbb{V}[X_d]}$ does not go to zero).

We note that the limitation of this model under this specific misspecification is mitigated by the inference objective in Eq. \ref{eq:altlb}, which also seeks out parameters useful for predicting the targets for a larger range of $p$. We show evidence for this in Supplement Sec. 3, but restrict our discussion and experiments to the maximum likelihood objective here. Exploring this objective is potentially interesting future work.

\section{Experimental Details}
\label{sec:experiments}

\begin{figure*}[ht]
    \centering
    \begin{subfigure}[t]{0.3\textwidth}
        \includegraphics[width=\textwidth]{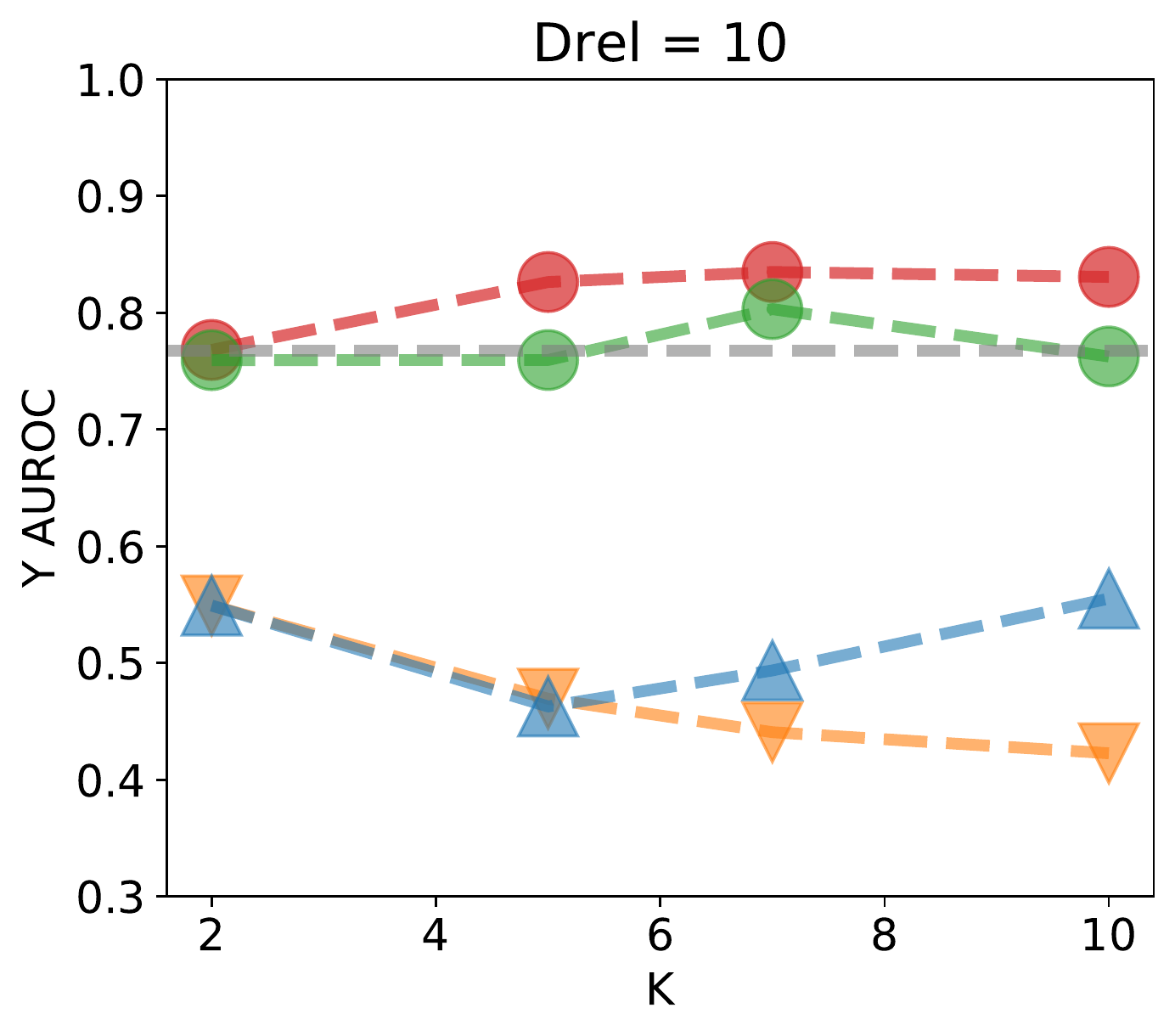}
        \caption{Synthetic}
        \label{fig:yroc2}
    \end{subfigure}
       \begin{subfigure}[t]{0.3\textwidth}
        \includegraphics[width=\textwidth]{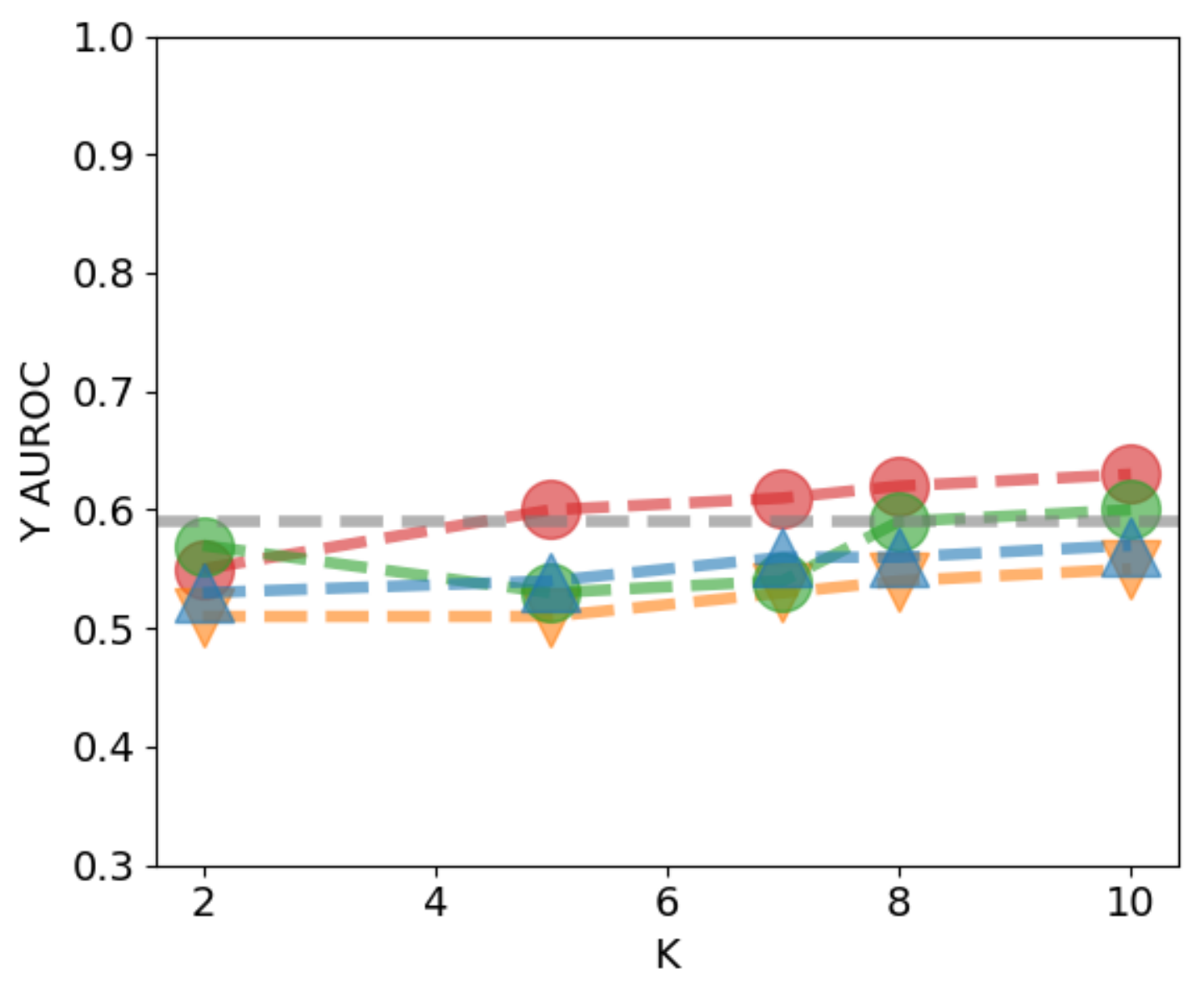}
        \caption{HIV}
        \label{fig:yroc1}
    \end{subfigure}
    \begin{subfigure}[t]{0.3\textwidth}
        \includegraphics[width=\textwidth]{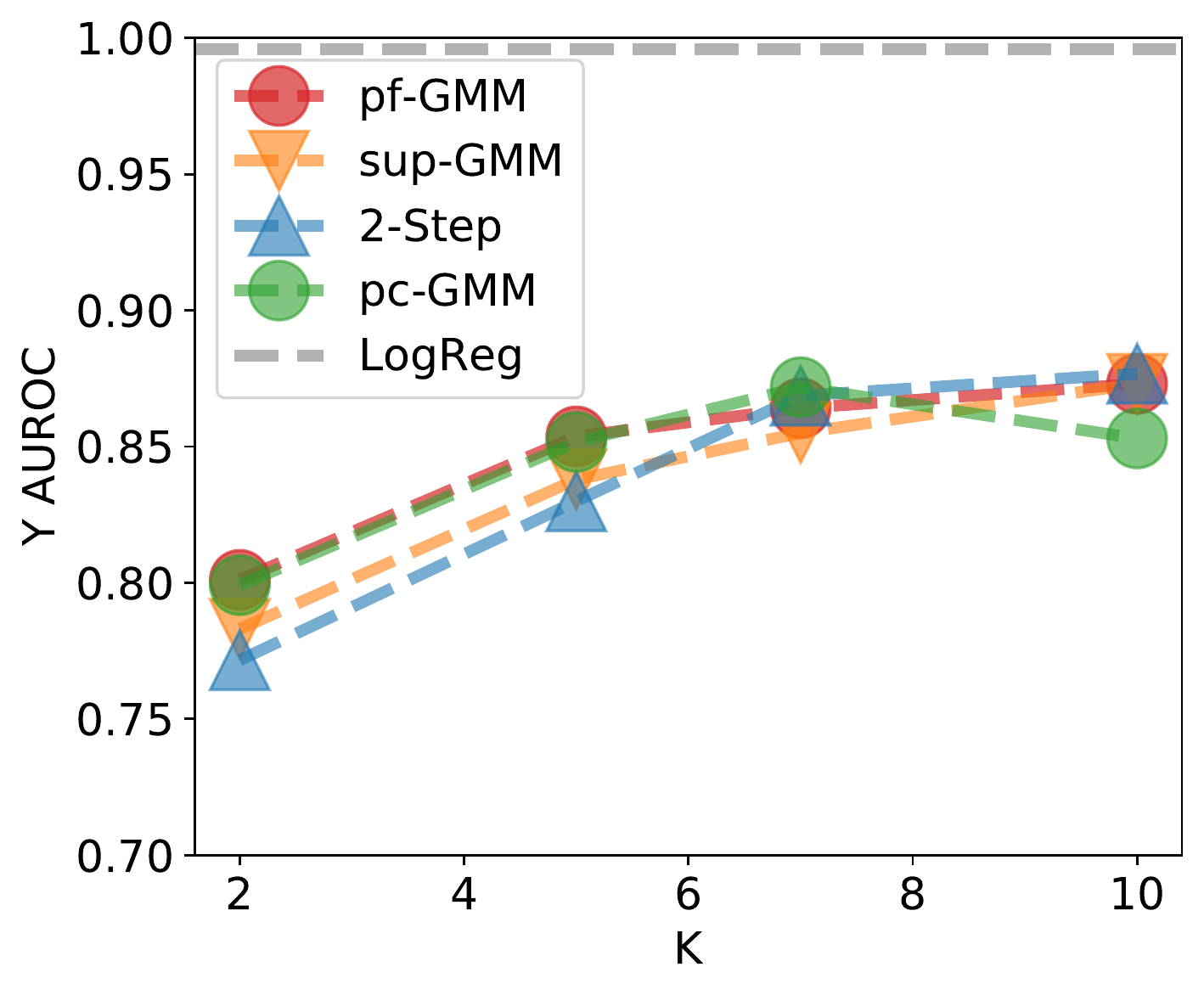}
        \caption{HAR}
        \label{fig:yroc3}
    \end{subfigure}
    \begin{subfigure}[t]{0.3\textwidth}
        \includegraphics[width=\textwidth]{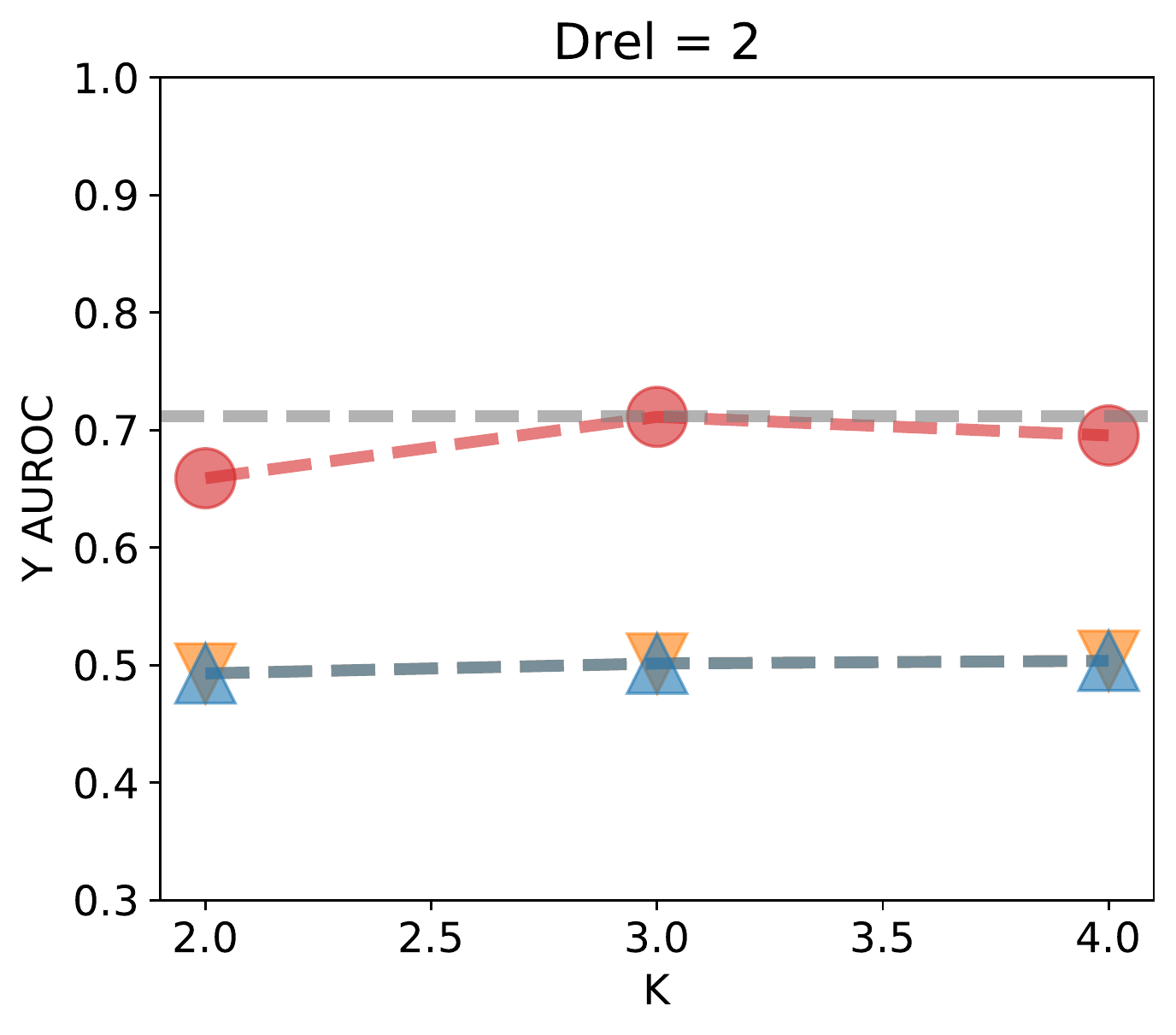}
        \caption{Synthetic}
        \label{fig:yroc5}
    \end{subfigure}
        \begin{subfigure}[t]{0.3\textwidth}
        \includegraphics[width=\textwidth]{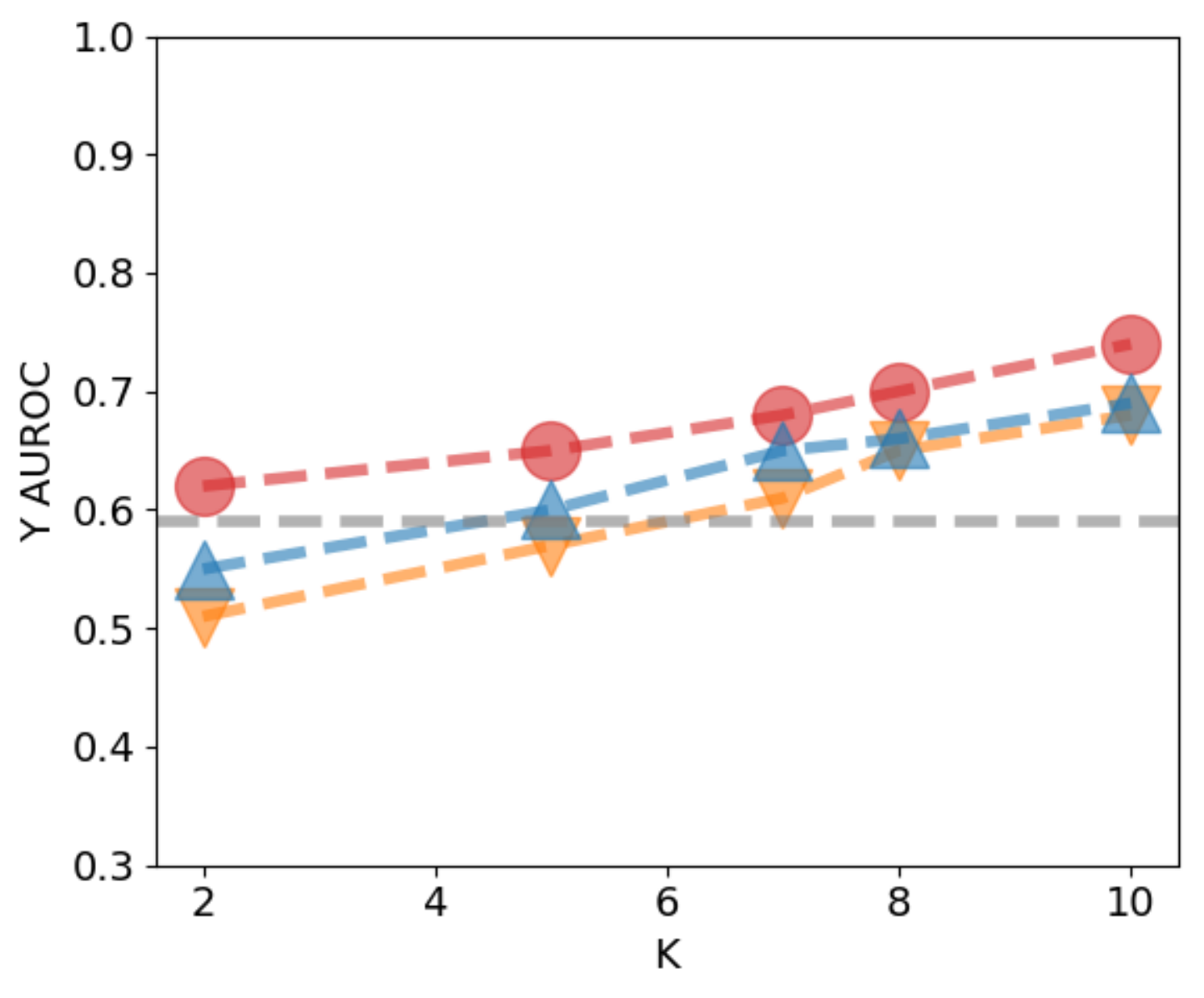}
        \caption{HIV}
        \label{fig:yroc4}
    \end{subfigure}
    \begin{subfigure}[t]{0.3\textwidth}
        \includegraphics[width=\textwidth]{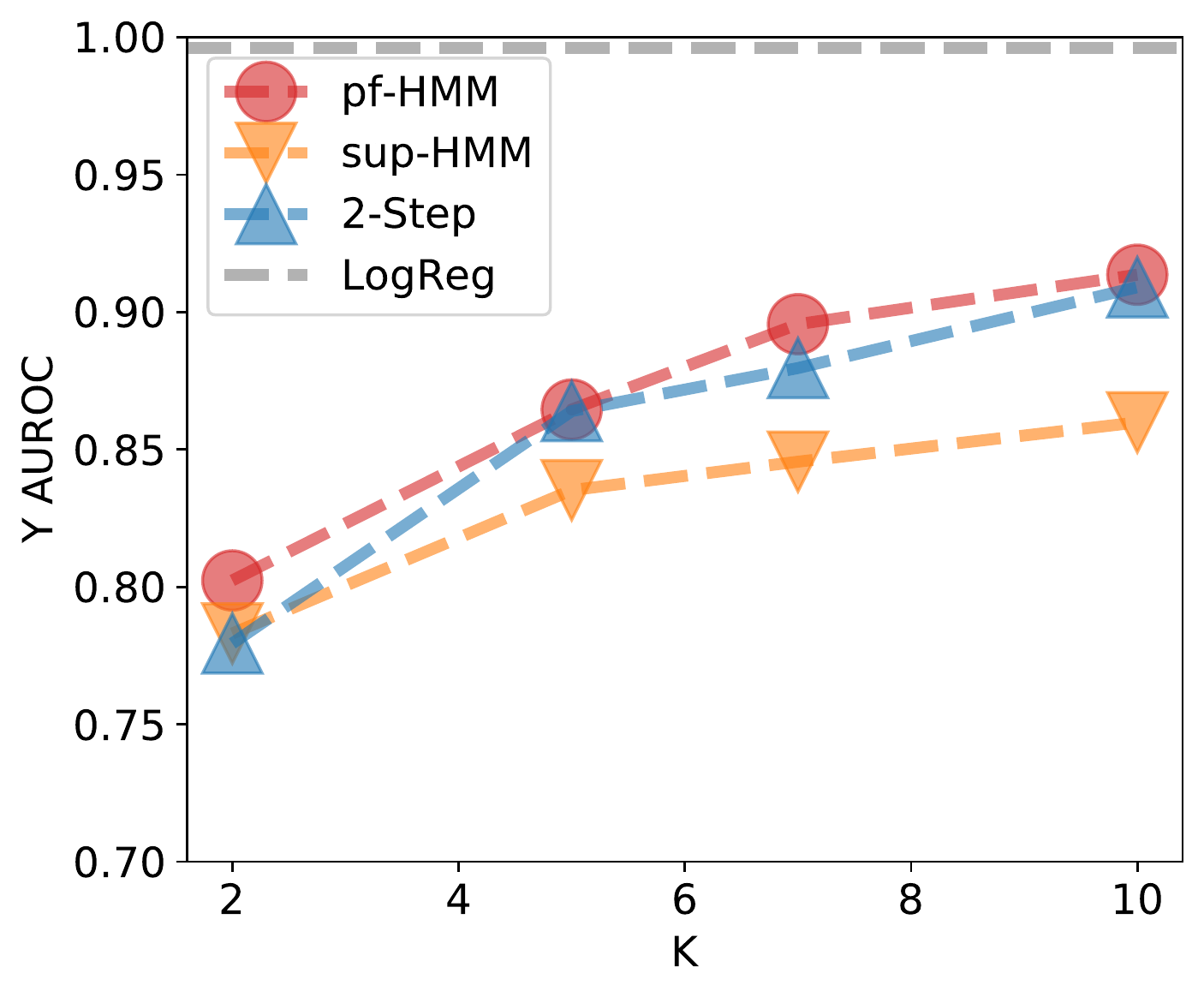}
        \caption{HAR}
        \label{fig:yroc6}
    \end{subfigure}
    \caption{Target AUROC performance as we vary the component budget $K$ in each of the domains for GMMs (Top Row) and HMMs (Bottom Row) respectively. \pfGMM s and \pfHMM s consistently outperform methods that do not perform the \textit{prediction-generation trade-off} for different choices of $K$. The synthetic input data for the \GMM\ case has 20 (out of 100 total) signal dimensions, with both signal and noise dimensions created using independent GMMs of 10 components each. For the \HMM\ case, the corresponding values are 2 (out of 20) and 4 respectively.}
    \label{fig:yroc_vs_k}
\end{figure*}

In this section, we demonstrate that the \pfGMM\ can select relevant signal from the input in both misspecified and fully specified settings. We demonstrate that the model structure---involving the switch parameters and their prior---is able to recover useful dimensions and maintain downstream performance even under misspecification.
\paragraph{Baselines.}
We compare the generative and predictive performance of \pfGMM\ to (1) an approach that first learns a generative model using only the inputs and then trains a logistic regression classifier on the posterior beliefs $p(Z | \xseq)$ (\twoGMM), (2) a supervised Gaussian mixture model without the switch parameters (\supGMM), (3) a logistic regression model, which is a  discriminative approach that maximizes $\log p(Y | \xseq)$ directly (\logreg), and (4) a prediction-constrained GMM \parencite{hughes2017predictionconstrained} (\pcGMM).
% \note{we are yet to run pc-GMM. also, i think it's better to skip logistic regression and instead have the discriminative term on \pfGMM\ baseline}
For the \pfHMM, we similarly compare to \supHMM, \twoHMM\ and \logreg.
\paragraph{Evaluation.}
To evaluate prediction quality, we compare the AUROC for the categorical target dimension on heldout data. To further assess models on their generative and discriminative quality, we compute $\log p(\xseq)$ and $\log p(\yseq | \xseq)$ on heldout data (computed without any approximations to ensure consistent evaluation between the methods).
% Due to the cost of computing \loglike\ and \altlb\ with the number dimensions as $\mathcal{O}(D^3)$, we restrict our comparison to these objectives only when $D = 10$.
\paragraph{Datasets.}
\emph{Synthetic.}
% Using experiments on synthetic datasets, we understand the behavior of mixture models on varying degrees of misspecification.
We use synthetic experiments to understand the behavior of mixture models of varying degrees of misspecification.
For our first experiment, we test whether \pfGMM\ is robust to increasing the number of noise dimensions. To simulate this setting, we vary the number of dimensions that come from the `signal' mixture while keeping the number of dimensions fixed at 100. Next, we test whether our model identifies the dimensions of interest under different budgets of components. Finally, we test whether the inferred posterior values for the switches indeed correspond to the true relevance of the dimensions.

We generate the signal and noise dimensions from independent GMMs, each with 10 components (we do the same for HMMs with 4 components). The emission distribution of each mixture given the cluster identity K is $p(\xseq | Z=K) = \Normal{X; 6K, I}$. The target is assigned a label in $\set{0,1}$, with the assignment only depending on the cluster identities in the signal mixture. The setup is presented in detail in Supplement Sec. 4.
\begin{figure*}[t]
    \centering
    \begin{subfigure}[t]{0.3\textwidth}
        \includegraphics[width=\textwidth]{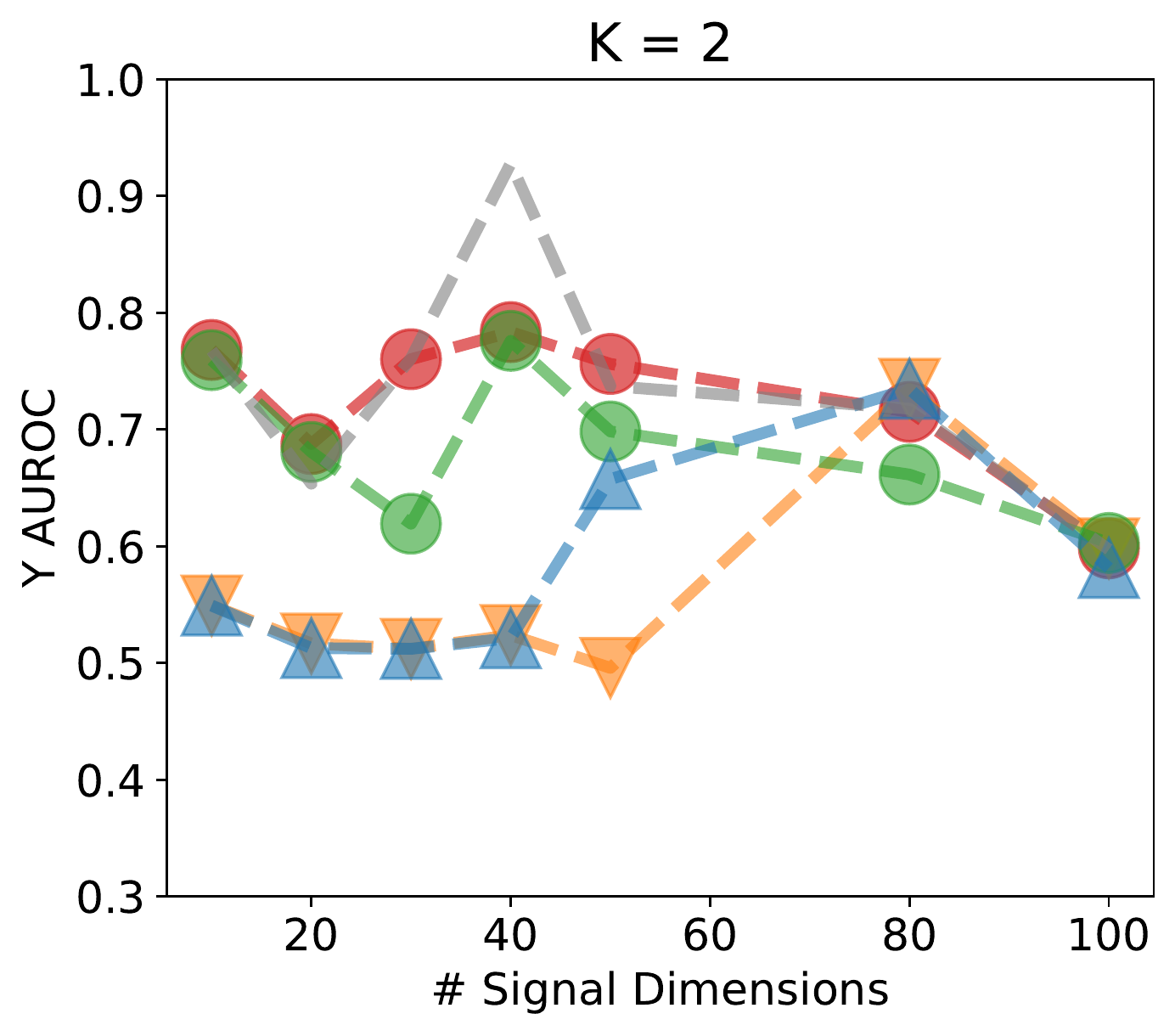}
        \caption{}
        \label{fig:yroc_vs_drel}
    \end{subfigure}
    \begin{subfigure}[t]{0.3\textwidth}
        \includegraphics[width=\textwidth]{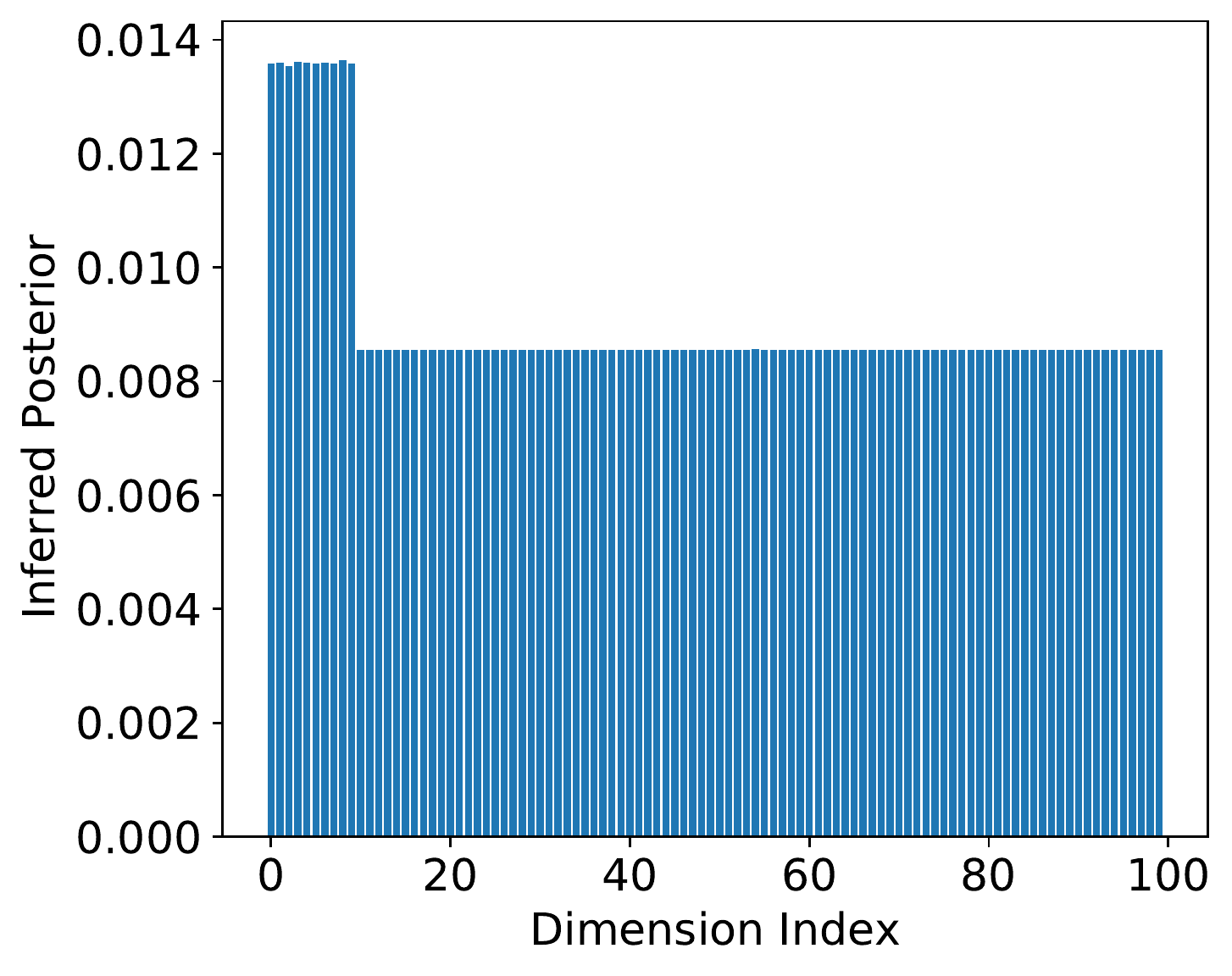}
        \caption{}
        \label{fig:phi_plot}
    \end{subfigure}
    \begin{subfigure}[t]{0.3\textwidth}
        \includegraphics[width=\textwidth]{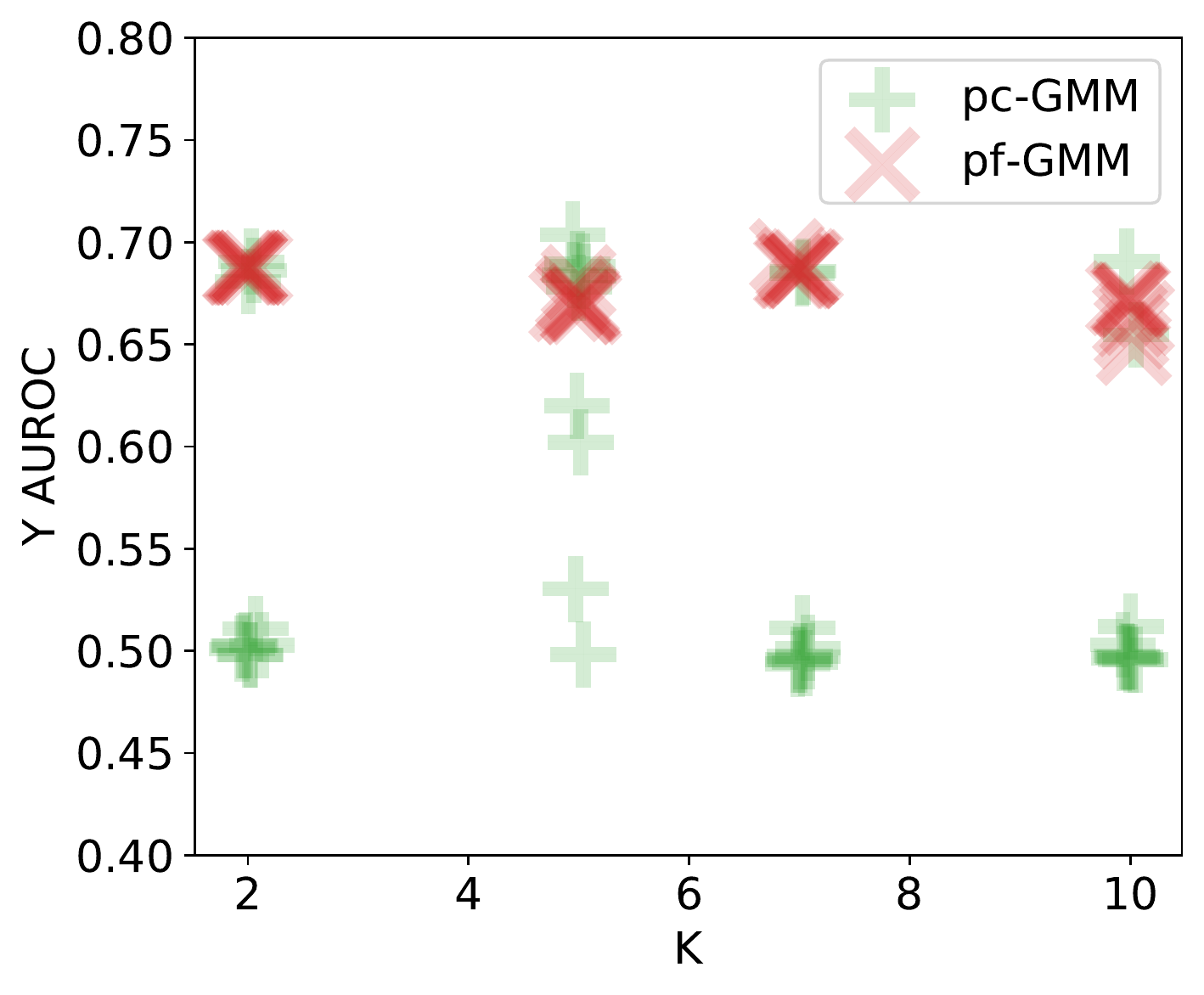}
        \caption{}
        \label{fig:pc_vs_pf_opt}
    \end{subfigure}
    \caption{For the above set of plots, signal and noise dimensions (100 in total) were created using 2 independent GMMs of 10 components each; our component modeling budget is 2. (a) Target AUROC as we vary the number of signal dimensions in the data. When noise dimensions outnumber signal, \pfGMM\ consistently outperforms methods which do not effectively trade off prediction and generation quality. (b) Inferred switch posterior distribution when first 10 dimensions are relevant in the ground truth; \pfGMM\ correctly learns to distinguish between signal and noise. (c) Variety in performance of optimal hyperparameters of \pcGMM\ and \pfHMM\ under different component budgets (20 signal dimensions in data).}
    \label{fig:one_example}
\end{figure*}

\emph{HIV.}
HIV is a virus that attacks the immune system which, if untreated, causes AIDS. Therapy for HIV involves administering cocktails of antiretrovirals from five classes namely, nnRTIs, nRTIs, PIs, FIs, and IIs to bring the viral load below detection limits ($\leq 40 $ copies/ml). Even in the presence of treatment, the virus may mutate. Some mutations may cause drug resistance and thus increase the viral load. These mutations are known as resistance-relevant mutations and can be identified from a list of mutations by the IAS USA \parencite{Gunthard2019}. We study 53\,236 patients with HIV from the EuResist Integrated Database \parencite{zazzi2012predicting}. Each person has a timeseries of approximately 16 steps associated with them. Our task is to predict whether a treatment will bring the viral load below detection limits in the next time-step. Each input contains 138 features including CD4$^{+}$ counts, mutations, treatments in terms of drug classes and lab results. We compare the performance of \pfGMM\ and \pfHMM\ with their corresponding baselines. Further details about the experimental setup are provided in the Supplement Sec. 4.

\emph{Human Activity Recognition Dataset.}
The Human Activity Recognition \parencite{Anguita_Ghio_Oneto_Parra_Reyes-Ortiz_2013} dataset uses a smartphone to capture triaxial acceleration, angular velocity, and other measurements of 30 participants while they perform one of 6 activities of daily living. Each person has one time series associated with them, with an average length of 338. Our task is to predict the daily activity the participant performs at each timestep. We compare the predictive performance of both \pfHMM\ and \pfGMM\ at different component budgets.
% \vspace{-0.5cm}

\section{Results}
\label{sec:results}
\paragraph{Prediction-focused models identify parameters which perform better at the downstream task.}
\pfGMM\ and \pfHMM\ outperform their vanilla counterparts on target AUROC on several real and synthetic datasets (Figures \ref{fig:analysis3} and \ref{fig:yroc_vs_k}).
This is because their objectives maximize the data likelihood while paying attention to the targets at training time. 
While the supervised variants also maximize the joint likelihood of inputs and target, they treat the target as just another input dimension. In contrast, pf-- learning identifies the asymmetry in the problem and puts the focus on input dimensions that are predictive of the targets.

\begin{figure}[t]
    \centering
    \includegraphics[width=0.45\textwidth]{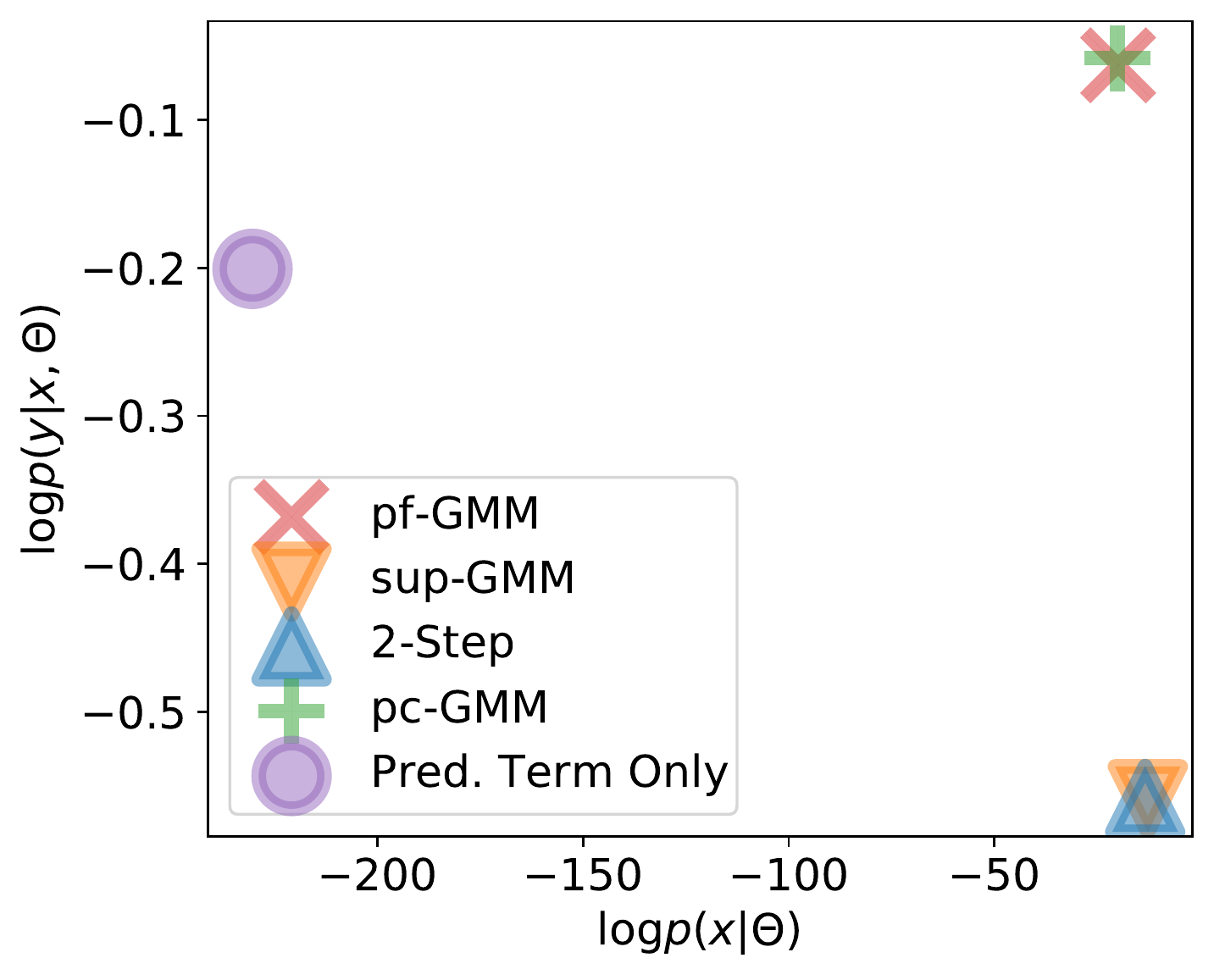}
    \caption{Discriminative-generative landscape. Here, 1 (out of 8) dimension is relevant; both signal and noise were generated from different GMMs with 2 components; our modeling budget is 2 components}
    \label{fig:disc_vs_gen}
\end{figure}

\paragraph{Prediction-focused models find relevant dimensions in misspecified and high-noise settings.}
Fig. \ref{fig:yroc_vs_drel} shows that the \pfGMM\ is able to make the best of its budget of two components even when it is severely constrained (both signal and noise GMMs had 10 components) and misspecified w.r.t the noise. This is possible because the model learns to focus on the dimensions that matter, as can be seen from its inferred switch posterior in Fig. \ref{fig:phi_plot}.
For HIV, we observe that for patients where the pf-- models predict that a treatment fails to bring the viral load below detection limits, a number of resistance-relevant mutations associated with that treatment are identified. 
As an example, for a patient taking nRTIs experiencing virologic failure, pf-- models identify certain mutations such as K65R and M184V as relevant. These are consistent with \cite{Gunthard2019}. Such insights may ultimately be useful for assessing treatment options as well as offering alternative options for intervention.
% \vspace{-0.3cm}
\paragraph{Prediction-focused models balance the predictive and generative objective.}
To demonstrate that pf-- models effectively trade off predictive and generative objectives, we plot its performance on the discriminative vs generative objective landscape (Fig. \ref{fig:disc_vs_gen}). 
Under severe misspecification, the \supGMM\ and \twoGMM\ models are not able to separate the signal from noise and thus pay a price in discriminative performance.
We also plot the \pfGMM\ model which is directly trained on the discriminative objective, $\log p(Y|\xseq)$, and see that focusing just on the discriminative term results in having poor generative quality. \pcGMM\ also does well with respect to the trade-off, because it explicitly incorporates $\log p(Y|\xseq)$ and $\log p(\xseq)$ in its objective.

\paragraph{The switch posteriors can be interpreted as usefulness of the dimension towards prediction.}
Figures \ref{fig:analysis5} and \ref{fig:phi_plot} demonstrate that the switch posterior identifies dimensions that were relevant for prediction. Importantly, this property is always true --- not just when pf-- models outperform their non-pf-- counterparts. None of the other methods assign an interpretable notion of relevance to the inputs.
% \vspace{-.3cm}
\paragraph{The switch prior is the only hyperparameter which we tune.}
The only hyperparameter, $p$, is intuitive to tune because increasing it monotonically increases our belief about how many relevant dimensions are present. This also means that setting $p=0$ skips using the inputs altogether and setting $p=1$ treats all dimensions as relevant. The optimal $p$ therefore ends up somewhere in between, as also noted in Fig. \ref{fig:analysis4}.
% \vspace{-.3cm}
\paragraph{Prediction-focused models are easier to optimize and achieve more stable solutions.}
pf-- modeling benefits from being a valid graphical model and thus allowing exact variational EM updates for the parameters. This means that our objective is non-decreasing after every step---which is true for the vanilla methods but not for \pcGMM. Because it has just one hyperparameter, searching over it is just a 1-D search. 
While \pcGMM\ has just one hyperparameter $\lambda$---used for trading off generative with predictive performance---it is still considerably harder to optimize. The optimization requires a grid search over both learning rates and the $\lambda$ hyperparameter, and takes much longer to converge than the EM-based methods. We see from Fig. \ref{fig:pc_vs_pf_opt}, that even after conditioning on the best hyperparameters for each setting, \pcGMM\ solutions are much less stable and would typically require more independent trials when compared to \pfGMM\ solutions.

\section{Discussion and Conclusion}
In this paper, we study mixture modeling under misspecification and propose a graphical model solution to the problem. This prediction-focused modeling selects the `right' features to cluster even when the inputs are corrupted with several disparate noise mixtures. It achieves this trade-off by maximizing the likelihood of the data directly, instead of resorting to any property of a modified objective as proposed by \textcite{hughes18-pclda} or of the approximate posterior as done by \textcite{ren2020prediction}. We characterize the conditions where this model outperforms its counterparts as well as the conditions where it does not. We also show that prediction-focused models perform at least as well as supervised mixture models and mixture models trained without any target, and therefore are a useful alternative to these widely used models.

% \subsubsection*{Acknowledgements}

% To preserve the anonymity, please include acknowledgments \emph{only} in the camera-ready papers.

% \subsubsection*{References}
% \defbibheading{subbibliography}[\refname]{\subsection*{#1}}
\printbibliography[heading=subbibliography]

@article{alemi2016deep,
  title   = {Deep variational information bottleneck},
  author  = {Alemi, Alexander A and Fischer, Ian and Dillon, Joshua V and Murphy, Kevin},
  journal = {arXiv preprint arXiv:1612.00410},
  year    = {2016}
}

@article{zazzi2012predicting,
  title={Predicting response to antiretroviral treatment by machine learning: the EuResist project},
  author={Zazzi, Maurizio and Incardona, Francesca and Rosen-Zvi, Michal and Prosperi, Mattia and Lengauer, Thomas and Altmann, Andre and Sonnerborg, Anders and Lavee, Tamar and Sch{\"u}lter, Eugen and Kaiser, Rolf},
  journal={Intervirology},
  volume={55},
  number={2},
  pages={123--127},
  year={2012},
  publisher={Karger Publishers}
}

@article{Gunthard2019,
  title={2019 update of the drug resistance mutations in HIV-1},
  author={Wensing, Annemarie M and Calvez, Vincent and Ceccherini-Silberstein, Francesca and Charpentier, Charlotte and G{\"u}nthard, Huldrych F and Paredes, Roger and Shafer, Robert W and Richman, Douglas D},
  journal={Topics in antiviral medicine},
  volume={27},
  number={3},
  pages={111},
  year={2019},
  publisher={International Antiviral Society--USA}
}

@article{blei_mcauliffe,
  author  = {David M. Blei and Jon D. McAuliffe},
  title   = {Supervised Topic Models},
  journal = {Proceedings of the 20th International Conference on Neural Information Processing Systems},
  year    = {2007},
  url     = {https://arxiv.org/pdf/1003.0783.pdf}
}

@article{cios2002uniqueness,
  title     = {Uniqueness of medical data mining},
  author    = {Cios, Krzysztof J and Moore, G William},
  journal   = {Artificial intelligence in medicine},
  volume    = {26},
  number    = {1-2},
  pages     = {1--24},
  year      = {2002},
  publisher = {Elsevier}
}

@inproceedings{ghahramani1994supervised,
  title     = {Supervised learning from incomplete data via an EM approach},
  author    = {Ghahramani, Zoubin and Jordan, Michael I},
  booktitle = {Advances in neural information processing systems},
  pages     = {120--127},
  year      = {1994}
}

@article{hughes2017predictionconstrained, title={Prediction-Constrained Training for Semi-Supervised Mixture and Topic Models}, url={http://arxiv.org/abs/1707.07341}, note={arXiv: 1707.07341}, journal={arXiv:1707.07341 [cs, stat]}, author={Hughes, Michael and Weiner, Leah and Hope, Gabriel and McCoy Jr., Thomas H. and Perlis, Roy H. and Sudderth, Erik B. and Doshi-Velez, Finale}, year={2017}, month={Jul} }

@inproceedings{hughes18-pclda,
  title     = {Semi-Supervised Prediction-Constrained Topic Models},
  author    = {Hughes, Michael and Hope, Gabriel and Weiner, Leah and McCoy, Thomas and Perlis, Roy and Sudderth, Erik and Doshi-Velez, Finale},
  booktitle = {Proceedings of the Twenty-First International Conference on Artificial Intelligence and Statistics},
  pages     = {1067--1076},
  year      = {2018},
  volume    = {84},
  series    = {Proceedings of Machine Learning Research},
  month     = {09--11 Apr},
  publisher = {PMLR}
}

@article{kingma2014semisupervised,
  title   = {Semi-supervised learning with deep generative models},
  author  = {Kingma, Diederik P and Rezende, Danilo J and Mohamed, Shakir and Welling, Max},
  journal = {arXiv preprint arXiv:1406.5298},
  year    = {2014}
}

@inproceedings{NigamMTM98,
  author    = {Nigam, Kamal and McCallum, Andrew and Thrun, Sebastian and Mitchell, Tom M.},
  booktitle = {AAAI/IAAI},
  editor    = {Mostow, Jack and Rich, Chuck},
  isbn      = {0-262-51098-7},
  pages     = {792-799},
  publisher = {AAAI Press / The MIT Press},
  title     = {Learning to Classify Text from Labeled and Unlabeled Documents.},
  year      = 1998
}

@article{popcorn,
  author  = {Joseph Futoma and Michael C. Hughes and Finale Doshi-Velez},
  title   = {POPCORN: Partially Observed Prediction Constrained Reinforcement Learning},
  journal = {Proceedings of the Twenty Third International Conference on Artificial Intelligence and Statistics},
  year    = {2020},
  url     = {https://arxiv.org/pdf/2001.04032.pdf}
}

@inproceedings{ren2020prediction,
  title        = {Prediction Focused Topic Models via Feature Selection},
  author       = {Ren, Jason and Kunes, Russell and Doshi-Velez, Finale},
  booktitle    = {International Conference on Artificial Intelligence and Statistics},
  pages        = {4420--4429},
  year         = {2020},
  organization = {PMLR}
}

@article{seymouretal2019sepsis,
  title   = {Derivation, Validation, and Potential Treatment Implications of Novel Clinical Phenotypes for Sepsis},
  volume  = {321},
  issn    = {0098-7484},
  doi     = {10.1001/jama.2019.5791},
  number  = {20},
  journal = {JAMA},
  author  = {Seymour, Christopher W. and Kennedy, Jason N. and Wang, Shu and Chang, Chung-Chou H. and Elliott, Corrine F. and Xu, Zhongying and Berry, Scott and Clermont, Gilles and Cooper, Gregory and Gomez, Hernando and et al.},
  year    = {2019},
  pages   = {2003–2017}
}

@article{tishby2000information,
  title   = {The information bottleneck method},
  author  = {Tishby, Naftali and Pereira, Fernando C and Bialek, William},
  journal = {arXiv preprint physics/0004057},
  year    = {2000}
}

@article{Wieczorek_2020,
  title     = {On the Difference between the Information Bottleneck and the Deep Information Bottleneck},
  volume    = {22},
  issn      = {1099-4300},
  url       = {http://dx.doi.org/10.3390/e22020131},
  doi       = {10.3390/e22020131},
  number    = {2},
  journal   = {Entropy},
  publisher = {MDPI AG},
  author    = {Wieczorek, Aleksander and Roth, Volker},
  year      = {2020},
  month     = {Jan},
  pages     = {131}
}

@inproceedings{Anguita_Ghio_Oneto_Parra_Reyes-Ortiz_2013, title={A Public Domain Dataset for Human Activity Recognition using Smartphones}, booktitle={ESANN}, author={Anguita, D. and Ghio, A. and Oneto, L. and Parra, X. and Reyes-Ortiz, J. L.}, year={2013} }

@inproceedings{Ganchev_Taskar_Gama_2008, title={Expectation Maximization and Posterior Constraints}, volume={20}, url={https://proceedings.neurips.cc/paper/2007/hash/73e5080f0f3804cb9cf470a8ce895dac-Abstract.html}, booktitle={Advances in Neural Information Processing Systems}, publisher={Curran Associates, Inc.}, author={Ganchev, Kuzman and Taskar, Ben and Gama, João}, year={2008} }

@inproceedings{Lacoste–Julien_Huszár_Ghahramani_2011, title={Approximate inference for the loss-calibrated Bayesian}, ISSN={1938-7228}, url={http://proceedings.mlr.press/v15/lacoste_julien11a.html}, booktitle={Proceedings of the Fourteenth International Conference on Artificial Intelligence and Statistics}, publisher={JMLR Workshop and Conference Proceedings}, author={Lacoste–Julien, Simon and Huszár, Ferenc and Ghahramani, Zoubin}, year={2011}, month={Jun}, pages={416–424} }

@article{Cobb_Roberts_Gal_2018, title={Loss-Calibrated Approximate Inference in Bayesian Neural Networks}, url={http://arxiv.org/abs/1805.03901}, note={arXiv: 1805.03901}, journal={arXiv:1805.03901 [cs, stat]}, author={Cobb, Adam D. and Roberts, Stephen J. and Gal, Yarin}, year={2018}, month={May} }

@inproceedings{Stoyanov_Ropson_Eisner_2011, title={Empirical Risk Minimization of Graphical Model Parameters Given Approximate Inference, Decoding, and Model Structure}, ISSN={1938-7228}, url={https://proceedings.mlr.press/v15/stoyanov11a.html}, booktitle={Proceedings of the Fourteenth International Conference on Artificial Intelligence and Statistics}, publisher={JMLR Workshop and Conference Proceedings}, author={Stoyanov, Veselin and Ropson, Alexander and Eisner, Jason}, year={2011}, month={Jun}, pages={725–733} }

\pagebreak
% \widetext
\begin{center}
\textbf{\Large Supplementary Materials}
\end{center}

% \title{On Learning Prediction-Focused Mixtures: % Supplementary Materials}

% \vspace*{-1.5cm} 

\newcommand{\dataobs}{{\textbf{x}, \textbf{y}}}
\newcommand{\xseqobs}{{\textbf{x}}}
\newcommand{\yseqobs}{{\textbf{y}}}

\section{Derivations}
In this section we provide the derivation of \elbo\ and coordinate ascent updates for our prediction-focused models. The derivations shown below will be for the \pfHMM, where $N$ is the number of sequences and $T_n$ signifies the length of sequence $n$. The results for pf-GMMs directly follow by setting $N$ as the number of data points and setting each $T_n$ as 1. The parameters of the model are $\Theta = \set{\theta, A, B, \pi, \eta}$
\subsection{ELBO derivation}
The lower bound for the log likelihood of one sequence with the posterior distribution $\fullposterior$ is:
\begin{align*}
   \log p(\xseq=\textbf{x}, \yseq=\textbf{y})
   &\geq \expectation{\miniposterior}{ \frac{\log p(\latentvars, \dataobs | \Theta, p )}{\fullposterior}} \\
   &= \expectation{\miniposterior}{\log p(\dataobs | \zseq, \xiseq, \Theta )} - \kldivergence{\fullposterior}{p(\zseq, \xiseq | \paramtrans, \priorz, p)} \\
   &= \expectation{q}{\log p(\xseqobs | \zseq, \xiseq, \paramEmitRelevant, \paramEmitIrrelevant )} 
   + \expectation{q}{\log p(\yseqobs | \zseq, \paramsY )}
   + \expectation{q}{\log p(\zseq | \priorz, \paramtrans)}
   + \expectation{q}{\log p(\xiseq | p)} \\
   & \quad - \expectation{q}{\log q(\xiseq | \varphi)}
   - \expectation{q}{\log q(\zseq | \dataobs, \varphi, \Theta)}
\end{align*}
where the individual terms are (define $q(Z_t | \dataobs, \Theta, \varphi) = q(Z_t)$) :
\begin{enumerate}
    \item 
    \parbox[t]{\textwidth}{\vspace{-2.9em}
    \begin{align*}
        \expectation{q}{\log p(\xseqobs | \zseq, \xiseq, \paramEmitRelevant, \paramEmitIrrelevant )} 
        &= \expectation{q(\xiseq | \varphi) q(\zseq | \dataobs, \Theta, \varphi)}{\sum_{n=1}^N \sum_{t=1}^{T_n} \sum_{d=1}^D {\xi_{t,d}^n} \log p_B (x_{td}^n | {Z_t}^n, \paramEmitRelevant) + \parens{1- {\xi_{t,d}^n}} \log p_\pi (x_{td}^n | \paramEmitIrrelevant) } \\
        &=  \sum_{n=1}^N \sum_{t=1}^{T_n} \sum_{d=1}^D  {\varphi_{d}}  \expectation{q(Z_t^n)}{\log p_B (x_{td}^n | {Z_t^n}, \paramEmitRelevant)}
        + \parens{1- {\varphi_{d}}} \log p_\pi (x_{td}^n | \paramEmitIrrelevant) \\
    \end{align*}}

    \item 
    \parbox[t]{\textwidth}{\vspace{-2.9em}
    \begin{align*}
        \expectation{q}{\log p(\yseqobs | \zseq, \xiseq, \paramEmitRelevant, \paramEmitIrrelevant )} 
       &= \expectation{q(\zseq | \dataobs, \Theta, \varphi)}{\sum_{n=1}^N \sum_{t=1}^{T_n} \log p(y_t^n | Z_t^n, \paramsY)} = \sum_{n=1}^N \sum_{t=1}^{T_n} \expectation{q(Z_t^n)}{\log p (y_t^n | Z_t^n, \paramsY)}
    \end{align*}}

    \item 
    \parbox[t]{\textwidth}{\vspace{-2.9em}
    \begin{align*}
        \expectation{q}{\log p(\zseq | \priorz, \paramtrans)} - \expectation{q}{\log q(\zseq | \dataobs, \varphi, \Theta)}
        &= \sum_{n=1}^N \expectation{q(Z_1^n)}{\log \frac{p(Z_1^n | \priorz)}{q(Z_1^n | \dataobs)}} + \sum_{n=1}^N \sum_{t=2}^{T_n} \expectation{q(Z_{t-1, t}^n | \dataobs)}{\log \frac{p(Z_t^n | Z_{t-1}^n, \paramtrans)}{q(Z_t^n | Z_{t-1}^n, \dataobs)}}
    \end{align*}}

    \item 
    \parbox[t]{\textwidth}{\vspace{-2.9em}
    \begin{align*}
        \expectation{q}{\log p(\xiseq | p)} - \expectation{q}{\log q(\xiseq | \varphi)}
        &= \expectation{q(\xiseq | \varphi)}{\sum_{n=1}^N \sum_{t=1}^{T_n} \sum_{d=1}^D \log \frac{p(\xi_{t,d}^n | p)}{q(\xi_{t,d}^n | \varphi_d)}} \\
        &= \sum_{n=1}^N \sum_{t=1}^{T_n} \sum_{d=1}^D \varphi_d \log \frac{p}{\varphi_d} + \parens{1 - \varphi_d} \log \frac{1-p}{1 - \varphi_d} \\
        % &= \sum_{t=1}^T \sum_{d=1}^D \varphi_d \log \varphi_d + \parens{1 - \varphi_d} \log \parens{1 - \varphi_d} - \varphi_d \log p - \parens{1 - \varphi_d} \log \parens{1 - p} \\
    \end{align*}}
\end{enumerate}

\subsection{Coordinate Ascent Updates}
 We have the ELBO as:
\begin{align*}
    L &= \sum_{t} \parens{\sum_{d=1}^D {\varphi_d} \expectation{q(Z_t | \dataobs)}{\log p_B (x_{td} | Z_t)}
    + \parens{1 - \varphi_d} \log p_\pi (x_{td})}
    + \expectation{q(Z_T | \dataobs)}{\sum_{t=0}^{T_n} \log p (y_t | Z_t, \paramsY)} \\
    & - \sum_d \expectation{q(\xiseq_d | \varphi_d)}{\log \frac{q(\xiseq_d | \varphi_d)}{p(\xiseq_d | p_d)}} 
    - \expectation{q(Z_0 | \dataobs)}{\log \frac{q(Z_0 | \dataobs)}{p(Z_0 | \priorz)}} 
    - \sum_t \expectation{q(Z_{t-1, t} | \dataobs)}{\log \frac{q(Z_t | Z_{t-1}, \dataobs)}{p(Z_t | Z_{t-1}, \paramtrans)}} \\
    &= \expectation{q}{\log p(\xseqobs | \zseq, \xiseq, \paramEmitRelevant, \paramEmitIrrelevant )} 
    + \expectation{q}{\log p(\yseqobs | \zseq, \paramsY )}
    + \expectation{q}{\log p(\zseq | \priorz, \paramtrans)}
    + \expectation{q}{\log p(\xiseq | p)} \\
    & \quad - \expectation{q}{\log q(\xiseq | \varphi)}
    - \expectation{q}{\log q(\zseq | \dataobs, \varphi, \Theta)}
\end{align*}

\subsubsection{$\varphi$ update}

Relevant terms:
\begin{align*}
    L_{\varphi} &= \expectation{q}{\log p(\xseqobs | \zseq, \xiseq, \paramEmitRelevant, \paramEmitIrrelevant )} 
    + \expectation{q}{\log p(\xiseq | p)}
    - \expectation{q}{\log q(\xiseq | \varphi)} \\
    &= \sum_{n=1}^N \sum_{t=1}^{T_n} \sum_{d=1}^D {\varphi_{d}}  \expectation{q(Z_t)}{\log p_B (x_{td}^n | {Z_t}, \paramEmitRelevant)}
    + \parens{1- {\varphi_{d}}} \log p_\pi (x_{td}^n | \paramEmitIrrelevant)
    + \varphi_d \log \frac{p}{\varphi_d} + \parens{1 - \varphi_d} \log \frac{1-p}{1 - \varphi_d}
 \end{align*}
 Gradient of the loss terms:
 \begin{align*}
    \nabla_{\varphi_{d'}} L_{\varphi} = \sum_{n=1}^N \sum_{t=1}^{T_n} \expectation{q(Z_t)}{\log p_B (x_{td'}^n | {Z_t}, \paramEmitRelevant)} - \log p_\pi (x_{td'}^n | \paramEmitIrrelevant) + \log \frac{p}{1-p} - \log \frac{\varphi_{d'}}{1-\varphi_{d'}}
 \end{align*}
Update:
\begin{align*}
    \varphi_d &\gets \sigma \parens{\log \frac{p}{1-p} + \frac{ \sum_{n=1}^N \sum_{t=1}^{T_n} \expectation{q(Z_t^n | \dataobs, \varphi)}{\log p_B (x_{td}^n | Z_t^n, \paramsEmitRelevant} - \log p_\pi (x_{td}^n | \paramsEmitIrrelevant)}{\sum_{n=1}^N T_n}}
\end{align*}

\subsubsection{$\paramEmitRelevant$ update}

\underline{$\paramEmitRelevantMean$} \\ 

Relevant terms:
\begin{align*}
    L_{\paramEmitRelevantMean} &= \sum_{n=1}^N \sum_{t=1}^{T_n} \sum_{d=1}^D {\varphi_{d}}  \expectation{q(Z_t^n)}{\log p_B (x_{td}^n | {Z_t^n}, \paramEmitRelevant)} \\
    &= \sum_{n=1}^N \sum_{t=1}^{T_n} \sum_{d=1}^D {\varphi_{d}}  \expectation{q(Z_t^n)}{ \frac{-1}{2 \paramEmitRelevantVar_{Z_t^n,d}} \parens{x_{td}^n - \paramEmitRelevantMean_{Z_t^n,d}}^2 + c } 
\end{align*}

Gradient of the loss terms:
\begin{align*}
    \nabla_{\paramEmitRelevantMean_{j,d'}} L_{\paramEmitRelevantMean} &= 
    \sum_{n=1}^N \sum_{t=1}^{T_n} {\varphi_{d'}} \nabla_{\paramEmitRelevantMean_{j,d'}} \sum_{k=1}^K q(Z_t^n = k) \cdot \frac{-1}{2 \paramEmitRelevantVar_{k,d'}} \parens{x_{td'}^n - \paramEmitRelevantMean_{k,d}}^2 \\
    &= \sum_{n=1}^N \sum_{t=1}^{T_n} {\varphi_{d'}} q(Z_t^n = j) \cdot \frac{\parens{x_{td'}^n - \paramEmitRelevantMean_{j,d'}}}{\paramEmitRelevantVar_{k,d'}}
\end{align*}

Update:
\begin{align*}
    \paramEmitRelevantMean_{j,d'} &\gets \frac{\sum_{n=1}^N \sum_{t=1}^{T_n} q(Z_t^n = j) x_{td'}^n}{\sum_{n=1}^N \sum_{t=1}^{T_n} q(Z_t^n = j) }
\end{align*}

\underline{$\paramEmitRelevantVar$}

Relevant terms:
\begin{align*}
    L_{\paramEmitRelevantVar}
    &= \sum_{n=1}^N \sum_{t=1}^{T_n} \sum_{d=1}^D {\varphi_{d}}  \expectation{q(Z_t^n)}{ - \frac{1}{2} \log  \paramEmitRelevantVar_{Z_t^n,d} -  \frac{1}{2 \paramEmitRelevantVar_{Z_t^n,d}} \parens{x_{td}^n - \paramEmitRelevantMean_{Z_t^n,d'}}^2 + c }
\end{align*}

Gradient of the loss terms:
\begin{align*}
    \nabla_{\paramEmitRelevantVar_{j,d'}} L_{\paramEmitRelevantVar}
    &= \sum_{n=1}^N \sum_{t=1}^{T_n} {\varphi_{d'}} q(Z_t^n = j) \parens{ - \frac{1}{2 \paramEmitRelevantVar_{j,d'}} +  \frac{1}{2 \parens{\paramEmitRelevantVar_{j,d'}}^2} \parens{x_{td'}^n - \paramEmitRelevantMean_{j,d'}}^2 }
\end{align*}

Update (if $\varphi_{d'} \neq 0$):
\begin{align*}
    \paramEmitRelevantVar_{j,d'} &\gets 
    \frac{\sum_{n=1}^N \sum_{t=1}^{T_n} q(Z_t^n = j) \parens{x_{td'}^n - \paramEmitRelevantMean_{j,d'}}^2}{\sum_{n=1}^N \sum_{t=1}^{T_n} q(Z_t^n = j)}
\end{align*}

\subsubsection{$\paramEmitIrrelevant$ update}
\label{subsec:pi_m_update}

\underline{$\paramEmitIrrelevantMean$} 

Relevant terms:
\begin{align*}
    L_{\paramEmitIrrelevantMean} &= \sum_{n=1}^N \sum_{t=1}^{T_n} \sum_{d=1}^D \parens{1- {\varphi_{d}}} \log p_\pi (x_{td} | \paramEmitIrrelevant) \\
    &= \sum_{n=1}^N \sum_{t=1}^{T_n} \sum_{d=1}^D \parens{1- {\varphi_{d}}}  \frac{-1}{2 \paramEmitIrrelevantVar_{d}} \parens{x_{td}^n - \paramEmitIrrelevantMean_{d}}^2 + c
\end{align*}

Gradient of the loss terms:
\begin{align*}
    \nabla_{\paramEmitIrrelevantMean_{d'}} L_{\paramEmitIrrelevantMean} 
    &= \sum_{n=1}^N \sum_{t=1}^{T_n} \parens{1-\varphi_{d'}} \cdot \frac{\parens{x_{td'}^n - \paramEmitIrrelevantMean_{d'}}}{\paramEmitIrrelevantVar_{d'}}
\end{align*}

Update (if $\varphi_{d'} \neq 1$):
\begin{align*}
    \paramEmitIrrelevantMean_{d'} &\gets \frac{\sum_{n=1}^N \sum_{t=1}^{T_n} x_{td'}^n}{\sum_{n=1}^N \sum_{t=1}^{T_n} 1} \\
    &= \frac{\sum_{n=1}^N \sum_{t=1}^{T_n} x_{td'}^n}{\sum_{n=1}^N T_n}
\end{align*}

\underline{$\paramEmitIrrelevantVar$}

Relevant terms:
\begin{align*}
    L_{\paramEmitIrrelevantVar}
    &= \sum_{n=1}^N \sum_{t=1}^{T_n} \sum_{d=1}^D \parens{1-\varphi_{d}}\parens{ - \frac{1}{2} \log  \paramEmitIrrelevantVar_{d} -  \frac{1}{2 \paramEmitIrrelevantVar_{d}} \parens{x_{td}^n - \paramEmitIrrelevantMean_{d'}}^2 + c }
\end{align*}

Gradient of the loss terms:
\begin{align*}
    \nabla_{\paramEmitIrrelevantVar_{d'}} L_{\paramEmitIrrelevantVar} 
    &= \sum_{n=1}^N \sum_{t=1}^{T_n} \parens{1-\varphi_{d'}} \parens{ - \frac{1}{2 \paramEmitIrrelevantVar_{d'}} +  \frac{1}{2 \parens{\paramEmitIrrelevantVar_{d'}}^2} \parens{x_{td'}^n - \paramEmitIrrelevantMean_{d'}}^2 }
\end{align*}

Update:
\begin{align*}
    \paramEmitIrrelevantVar_{d'} &\gets \frac{\sum_{n=1}^N \sum_{t=1}^{T_n} \parens{x_{td'}^n - \paramEmitIrrelevantMean_{d'}}^2 }{\sum_{n=1}^N \sum_{t=1}^{T_n} 1} \\
    &= \frac{\sum_{n=1}^N \sum_{t=1}^{T_n} {\parens{x_{td'}^n}^2} }{\sum_{n=1}^N T_n} - \parens{\paramEmitIrrelevantMean}^2
\end{align*}

\subsubsection{$\priorz$ update}

Relevant terms (including the constraint and the regularizer):
\begin{align*}
    L_{\priorz} &= \log p(\priorz | \alpha) + \braces{\sum_{n=1}^N \sum_{k=1}^K q(Z_1^n =k) {\log p(Z_1^n =k | \priorz)}} - \lambda_{\priorz} \left( \sum_{k=1}^K \priorz_k - 1\right)
\end{align*}

Gradient of the loss terms:
\begin{align*}
    \nabla_{\priorz_{k'}} L_{\priorz} &= -\lambda_{\priorz} + (\alpha - 1) \frac{1}{\priorz_{k'}} + \sum_{n=1}^N q(Z_1^n =k') \frac{1}{\priorz_{k'}} \\
    \nabla_{\lambda_{\priorz}} L_{\priorz} &= - \sum_{k=1}^K \priorz_k
\end{align*}

Update:
\begin{align*}
    \priorz_{k'} &\gets \frac{\alpha - 1 + \sum_{n=1}^N q(Z_1^n =k')}{K \alpha - K + N}\\
    &= \frac{\alpha - 1 + \expectation{}{N^1_{k'}}}{K \alpha - K + N}
\end{align*}

\subsubsection{$\paramtrans$ update}

Relevant terms:
\begin{align*}
    L_{\paramtrans} &= \braces{\sum_{n=1}^N \sum_{t=2}^{T_n} \sum_{j=1}^J \sum_{k=1}^K q(Z_{t-1}^n=j, Z_t^n=k) {\log p(Z_t^n = k | Z_{t-1}^n = j, \paramtrans)}} - \sum_{l=1}^K \lambda_{\paramtrans_l} (\sum_{k=1}^K \paramtrans_{lk} - 1)
\end{align*}

Gradient of the loss terms:
\begin{align*}
    \nabla_{\paramtrans_{j', k'}} L_{\paramtrans} &= -\lambda_{\paramtrans_{j'}} + \sum_{n=1}^N \sum_{t=2}^{T_n} q(Z_{t-1}^n=j', Z_t^n=k') \frac{1}{\paramtrans_{j',k'}} \\
    \nabla_{\lambda_{\paramtrans_{l'}}} L_{\paramtrans} &= - \sum_{k=1}^K \paramtrans_{l'k}
\end{align*}

Update:
\begin{align*}
    \paramtrans_{j', k'} &\gets \frac{\sum_{n=1}^N \sum_{t=2}^{T_n} q(Z_{t-1}^n=j', Z_t^n=k')}{\sum_{n=1}^N \sum_{t=2}^{T_n} \sum_{k=1}^K q(Z_{t-1}^n=j', Z_t^n=k)} 
    = \frac{\sum_{n=1}^N \sum_{t=2}^{T_n} q(Z_{t-1}^n=j', Z_t^n=k')}{\sum_{n=1}^N \sum_{t=2}^{T_n} q(Z_{t-1}^n=j')} \\
    &= \frac{\expectation{}{N_{j',k'}}}{\sum_{k'=1}^K \expectation{}{N_{j',k'}}}
\end{align*}

\section{Proofs}

In this section, we provide the proof for the limitation result in Section 6 of the main paper. Specifically, we prove that maximization of the joint likelihood $\log p(\data)$ fails to give predictive components when the irrelevant dimensions cannot be modeled \textit{well} under the noise distribution, $\pi$.

\newcommand{\trueposttheta}[1]{p(\latentvars | \data; {#1})}
\newcommand{\truepost}{\trueposttheta{\Theta}}
\newcommand{\pxiz}{q_{\xi_d, Z_n}}
\newcommand{\pxid}[1]{q_{\xi_{#1}}}
\newcommand{\pxi}{\pxid{d}}
\newcommand{\pz}{q_{Z_n}}
\newcommand{\pzgivenxid}[1]{q_{Z_n | \xi_{#1} }}
\newcommand{\pzgivenxi}{\pzgivenxid{d}}
\newcommand{\one}{{(1)}}
\newcommand{\two}{{(2)}}
\newcommand{\cpi}{{\color{red} \pi}}
\newcommand{\cB}{{\color{blue} B}}

\paragraph{Proof}
We have that the \elbo\ is equal to the log likelihood (for any parameters $\Theta$) when $KL(q(\latentvars)||\truepost)$ is zero:
\begin{align*}
    \log p(\data ; \Theta) &= \expectation{\truepost}{\log p(\data, \latentvars; \Theta)} + \mathbb{H}\brackets{\truepost}
\end{align*}

Since the entropy term is bounded, we closely look at the first term:

\begin{align*}
&\expectation{\truepost}{\log p(\data, \latentvars; \Theta)} \\
    &= \sum_d \expectation{\pxi}{\log p(\xi_d; p)} + \sum_n \expectation{\pz}{\log p(Z_n; \theta)} + \sum_n \expectation{\pz}{\log p(Y_n | Z_n; \eta)} \\
    &\quad + \sum_n \sum_d \expectation{\pxiz}{\log p(X_{nd} | \xi_d, Z_n; \Theta)} \\
    &= - \sum_d \mathbb{H}\brackets{p(\xi_d; p)} - \sum_n \mathbb{H}\brackets{p(Z_n; \theta)} + \sum_n \expectation{\pz}{\log p(Y_n | Z_n; \eta)} \\
    &\quad + \sum_n \sum_d \expectation{\pxiz}{\log p(X_{nd} | \xi_d, Z_n; \Theta)}
\end{align*}
where we use the short-hands: $\pxi:= p(\xi_d |\data; \Theta), \pz:= p(Z_n |\data; \Theta), \pxiz:= p(\xi_d, Z_n |\data; \Theta), \pzgivenxi:= p(Z_n | \xi_d, \data; \Theta)$. Again, we will generally have the first three terms bounded, we expand the last term:

\begin{align*}
& \sum_n \sum_d \expectation{\pxiz}{\log p(X_{nd} | \xi_d, Z_n; \Theta)} \\
    &= \sum_n \sum_d \expectation{\pxiz}{\xi_d \log p(X_{nd} | \xi_d=1, Z_n; B) + (1-\xi_d) \log p(X_{nd} | \xi_d=0; \pi)} \\
    &= \sum_n \sum_d \expectation{\pxiz}{\xi_d \log p(X_{nd} | Z_n; B_d) + (1-\xi_d) \log p(X_{nd}; \pi_d)} \\
    &= \sum_n \sum_d \log p(X_{nd}; \pi_d) + \sum_n \sum_d \expectation{\pxiz}{\xi_d \parens{\log p(X_{nd} | Z_n; B_d) - \log p(X_{nd}; \pi_d)}}
\end{align*}

\begin{align}
    &= \sum_n \sum_d \log p(X_{nd}; \pi_d) + \sum_n \sum_d \expectation{\pxi}{\xi_d \parens{\expectation{\pzgivenxi}{\log p(X_{nd} | Z_n; B_d) }- \log p(X_{nd}; \pi_d)}}
    \label{eq:lastterm}
    % &= \sum_n \sum_d \log p(X_{nd}; \pi_d) + \sum_n \sum_d \pxi \parens{\expectation{q(Z_n | \xi_d=1)}{\log p(X_{nd} | Z_n; B_d) }- \log p(X_{nd}; \pi_d)} 
\end{align}

Now consider the two local optima where we either pick out the components along $X_1$ or along $\xseq_{2:6}$. Call them $\Theta^\one = \set{B^\one, \pi^\one, \theta^\one}$ and $\Theta^\two = \set{B^\two, \pi^\two, \theta^\two}$ respectively.

For $B^\one$, the learned $B$ components would then be $\Normal{0, 1}$ and $\Normal{6, 1}$ distributions for the first dimension and be the same as $\pi^\one_{2:6}$ for the remaining dimensions.

For $B^\two$, the learned $B$ distribution would then be $\Normal{\textbf{0}, I_5}$ and $\Normal{\muvec, I_5}$ distributions for dimensions 2-6, and the same as $\pi^\two_1$ for the first dimension.

Also note that $\pi^\one = \pi^\two$ because they are learned to just be the best-fit univariate Gaussian per dimension (as derived in subsection \ref{subsec:pi_m_update}).

Then for $\Theta^\one$, the Eq. \ref{eq:lastterm} becomes:

\begin{align*}
     &= \sum_n \sum_d \log p(X_{nd}; \cpi_d^\one) + \sum_n \sum_d \expectation{\pxi^\one}{\xi_d \parens{\expectation{\pzgivenxi^\one}{\log p(X_{nd} | Z_n; \cB_d^\one) }- \log p(X_{nd}; \cpi_d^\one)}}\\
    &= \sum_n \sum_d \log p(X_{nd}; \cpi_d^\one) + \sum_n \expectation{\pxid{1}^\one}{\xi_1 \parens{\expectation{\pzgivenxid{1}^\one}{\log p(X_{n1} | Z_n; \cB_1^\one) }- \log p(X_{n1}; \cpi_1^\one)}}\\
    &\quad + \sum_{d=2}^D \expectation{\pxi^\one}{\xi_d \parens{\expectation{\pzgivenxi^\one}{\log p(X_{nd} | Z_n; \cB_d^\one) }- \log p(X_{nd}; \cpi_d^\one)}}
\end{align*}

\begin{align*}
    &= \sum_n \sum_d \log p(X_{nd}; \cpi_d^\one) + \sum_n \expectation{\pxid{1}^\one}{\xi_1 \parens{\expectation{\pzgivenxid{1}^\one}{\log p(X_{n1} | Z_n; \cB_1^\one) }- \log p(X_{n1}; \cpi_1^\one)}}\\
    &\quad + \sum_{d=2}^D \expectation{\pxi^\one}{\xi_d \parens{\expectation{\pzgivenxi^\one}{\log p(X_{nd} ; \cpi_d^\one) }- \log p(X_{nd}; \cpi_d^\one)}}
\end{align*}
which finally reduces to:
\begin{align}
    &= \sum_n \sum_d \log p(X_{nd}; \cpi_d^\one) + \sum_n \expectation{\pxid{1}^\one}{\xi_1 \parens{\expectation{\pzgivenxid{1}^\one}{\log p(X_{n1} | Z_n; \cB_1^\one) }- \log p(X_{n1}; \cpi_1^\one)}} \label{eq:lasttermtheta1}
\end{align}

Similarly for $\Theta^2$, the Eq. \ref{eq:lastterm} becomes:
\begin{align*}
    &= \sum_n \sum_d \log p(X_{nd}; \cpi_d^\two) + \sum_n \expectation{\pxid{1}^\two}{\xi_1 \parens{\expectation{\pzgivenxid{1}^\two}{\log p(X_{n1} | Z_n; \cB_1^\two) }- \log p(X_{n1}; \cpi_1^\two)}}\\
    &\quad + \sum_{d=2}^D \expectation{\pxi^\two}{\xi_d \parens{\expectation{\pzgivenxi^\two}{\log p(X_{nd} | Z_n; \cB_d^\two) }- \log p(X_{nd}; \cpi_d^\two)}}\\
    &= \sum_n \sum_d \log p(X_{nd}; \cpi_d^\two)
    + \sum_n \expectation{\pxid{1}^\two}{\xi_1 \parens{\expectation{\pzgivenxid{1}^\two}{\log p(X_{n1} ; \cpi_1^\two) }- \log p(X_{n1}; \cpi_1^\two)}}\\
    &\quad + \sum_{d=2}^D \expectation{\pxi^\two}{\xi_d \parens{\expectation{\pzgivenxi^\two}{\log p(X_{nd} | Z_n; \cB_d^\two) }- \log p(X_{nd}; \cpi_d^\two)}}\\
\end{align*}
which reduces to:
\begin{align}
    &= \sum_n \sum_d \log p(X_{nd}; \cpi_d^\two)
    + \sum_{d=2}^D \expectation{\pxi^\two}{\xi_d \parens{\expectation{\pzgivenxi^\two}{\log p(X_{nd} | Z_n; \cB_d^\two) }- \log p(X_{nd}; \cpi_d^\two)}} \label{eq:lasttermtheta2}
\end{align}

Using Equations \ref{eq:lasttermtheta1} and \ref{eq:lasttermtheta2}, we can write log likelihood ratio:
\begin{align*}
    \log p(\data ; \Theta^\one) - \log p(\data ; \Theta^\two)
  &= \color{blue} \mathbb{H}\brackets{\trueposttheta{\Theta^\one}} - \mathbb{H}\brackets{\trueposttheta{\Theta^\two}} \\
  &\quad + \color{blue} \sum_n \mathbb{H}\brackets{p(Z_n; \theta^\two)} - \sum_n \mathbb{H}\brackets{p(Z_n; \theta^\one)} \\
  &\quad + \color{blue} \sum_n \expectation{\pz^\one}{\log p(Y_n | Z_n; \eta^\one)} - \sum_n \expectation{\pz^\two}{\log p(Y_n | Z_n; \eta^\two)} \\
  &\quad + \color{purple} \sum_n \expectation{\pxid{1}^\one}{\xi_1 \parens{\expectation{\pzgivenxid{1}^\one}{\log p(X_{n1} | Z_n; B_1^\one) }- \log p(X_{n1}; \pi_1^\one)}} \\
  &\quad - \sum_{d=2}^D \expectation{\pxi^\two}{\xi_d \parens{\expectation{\pzgivenxi^\two}{\log p(X_{nd} | Z_n; B_d^\two) }- \log p(X_{nd}; \pi_d^\two)}}
\end{align*}
\begin{align*}
  &= \color{blue} O(1) \\
  &\quad + \color{purple} \sum_n \expectation{\pxid{1}^\one}{\xi_1 \parens{\expectation{\pzgivenxid{1}^\one}{\log p(X_{n1} | Z_n; B_1^\one) }- \log p(X_{n1}; \pi_1^\one)}} \\
  &\quad - \sum_{d=2}^D \expectation{\pxi^\two}{\xi_d \expectation{\pzgivenxi^\two}{\log p(X_{nd} | Z_n; B_d^\two) }} + \sum_{d=2}^D \expectation{\pxi^\two}{\xi_d \log p(X_{nd}; \pi_d^\two)} \\
  &= O(1) + \sum_{d=2}^D \expectation{\pxi^\two}{\xi_d \log p(X_{nd}; \pi_d^\two)} \\
  &\rightarrow_{\mu \rightarrow \infty} -\infty
\end{align*}
which implies that the likelihood ratio goes to 0. Therefore, $\Theta^\one$ will not be selected over $\Theta^\two$ as $\mu \rightarrow \infty$.

% \textit{In this section, we present the detailed proof of Lemma 3 and then [ ... ]}

\newpage
\section{Additional Experiments}

\subsection{Plot of the bound in Eq 3 in the main paper}

\begin{figure}[H]
    \centering
    \includegraphics[width=.6\textwidth]{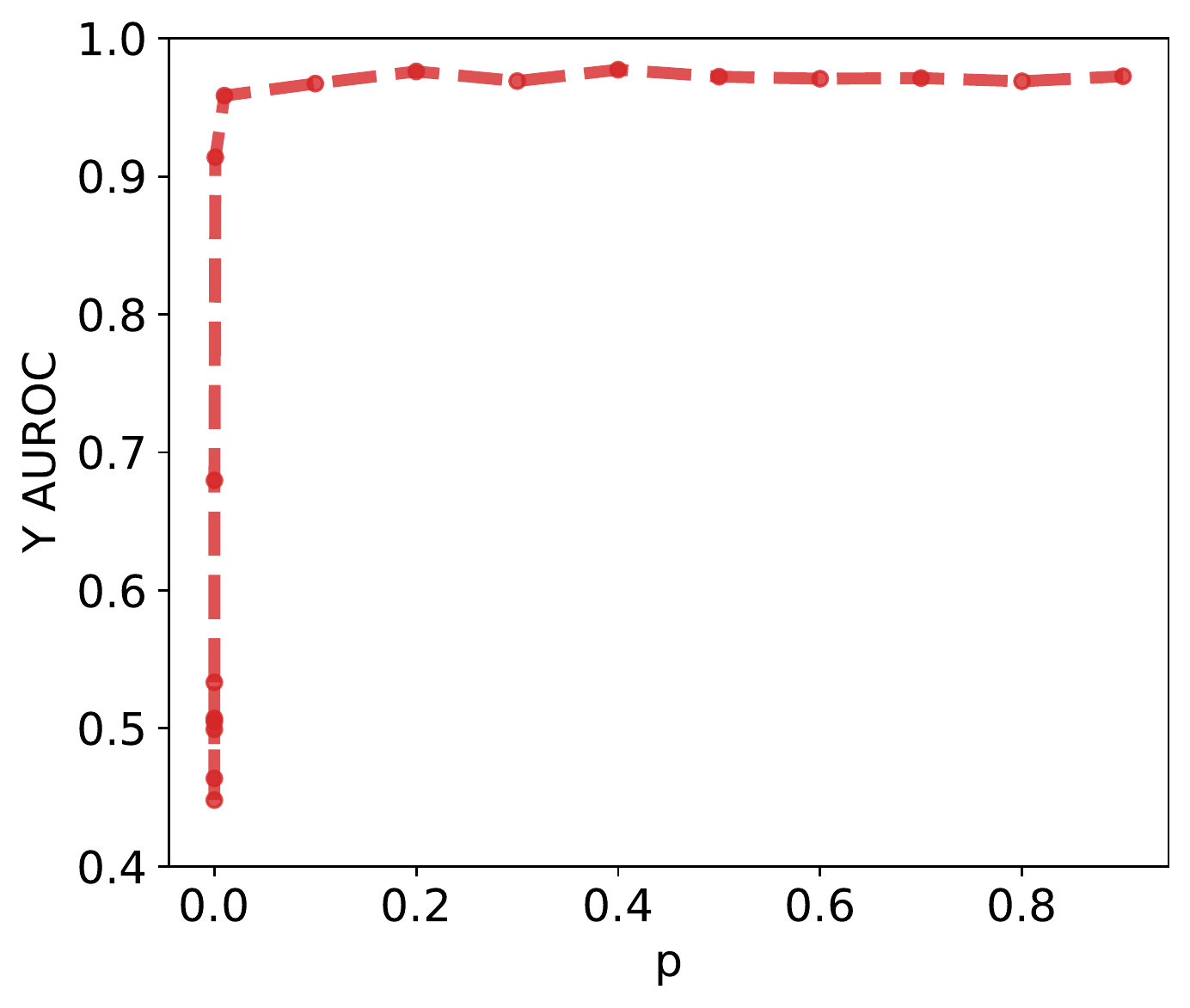}
    \caption{Heldout AUROC vs \pfGMM\ hyperparameter $p$ when we optimize the bound in Eq. 3 (main paper) instead of the likelihood---performance peaks for a much larger range of p, giving it better task-focused properties over a larger range of $p$. This would motivate looking for a tractable proxy for this objective (which is left to future work).}
    \label{fig:altlb_vs_p}
\end{figure}

\subsection{\pfGMM\ Simulated Data Experiments}
\begin{figure}[H]
\centering
\includegraphics[width=.3\textwidth]{figures/pfgmm_roc_vs_K_D_100_Drel_10.pdf}
\includegraphics[width=.3\textwidth]{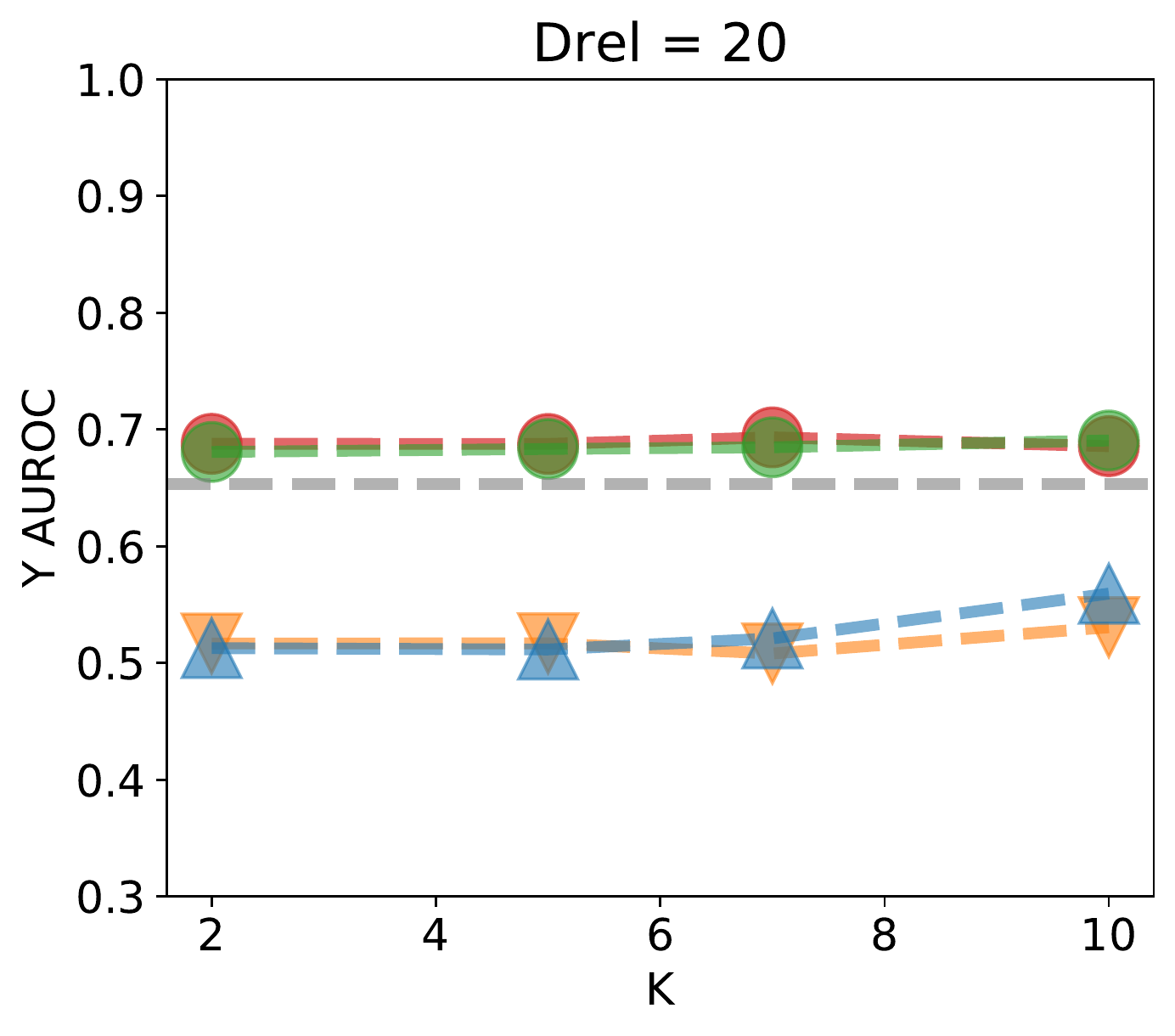}
\includegraphics[width=.3\textwidth]{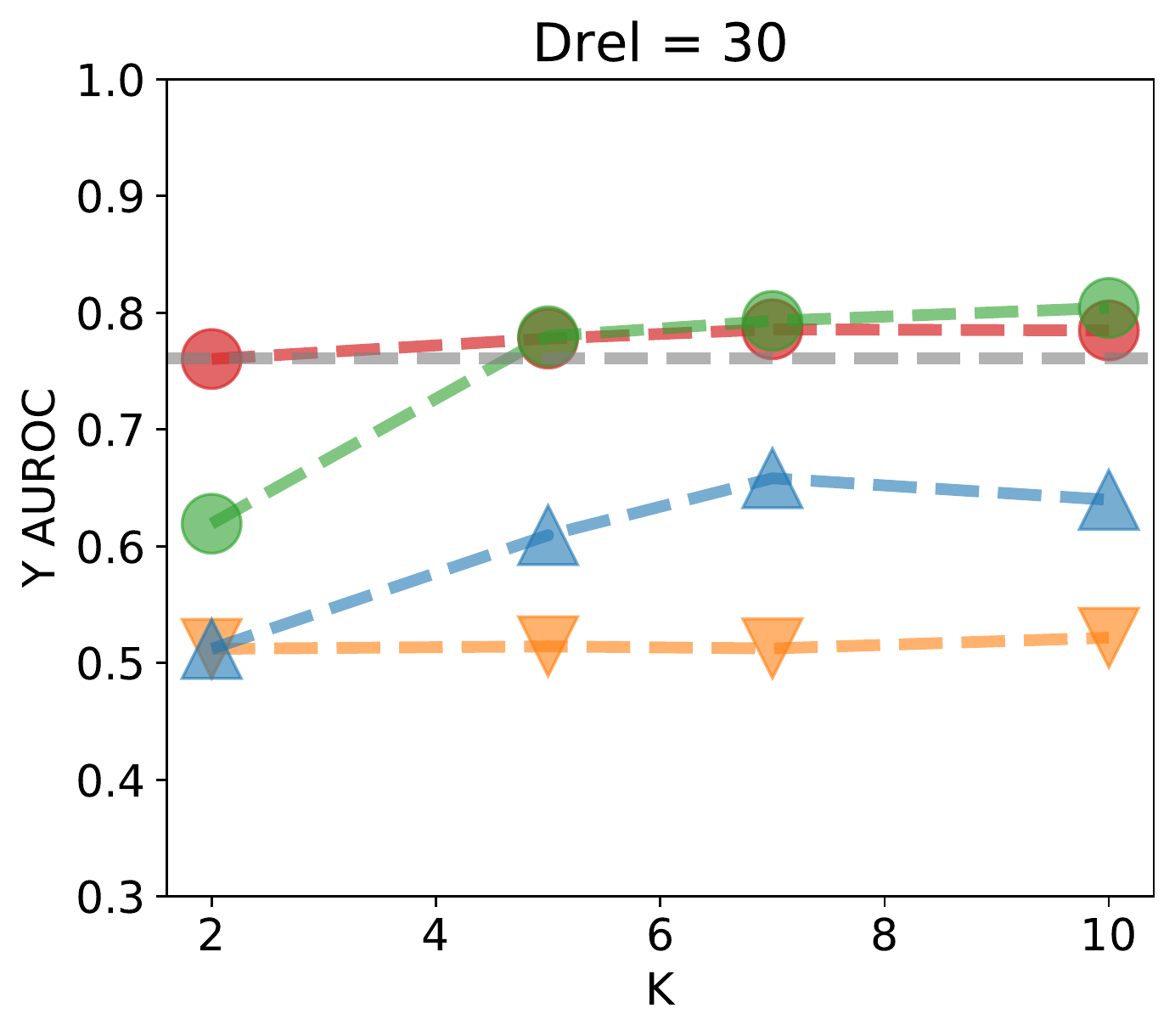}
\includegraphics[width=.3\textwidth]{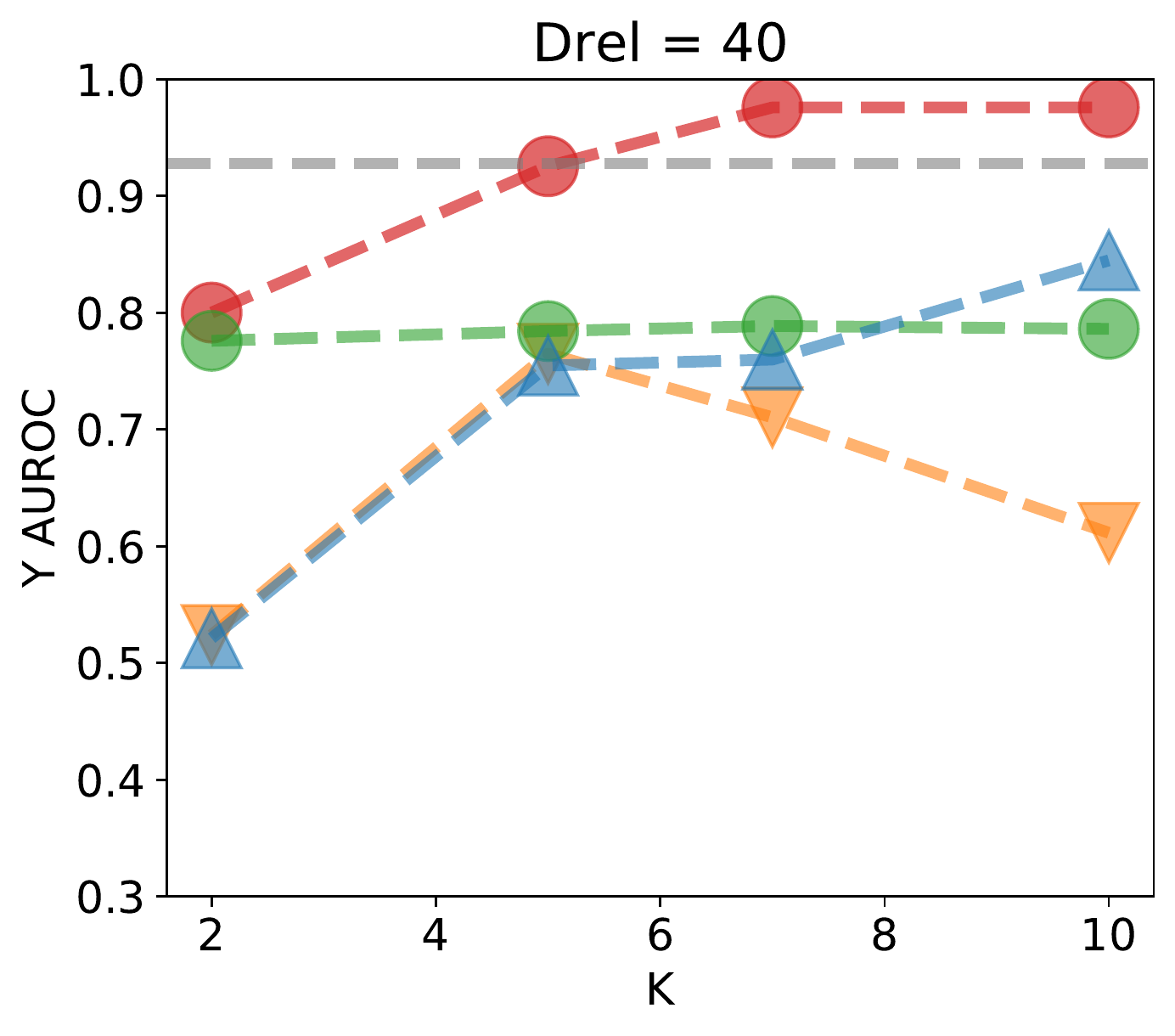}
\includegraphics[width=.3\textwidth]{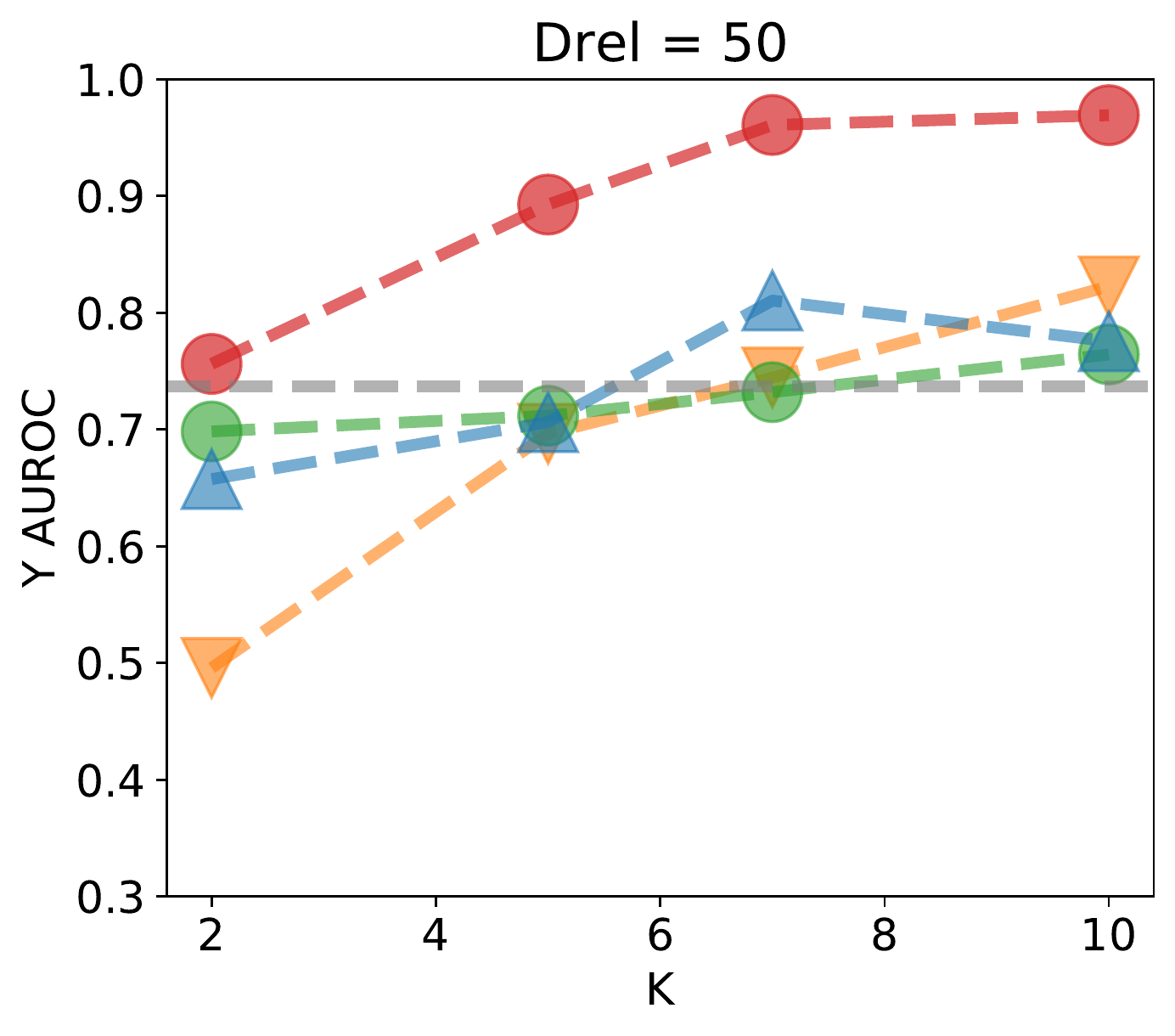}
\includegraphics[width=.3\textwidth]{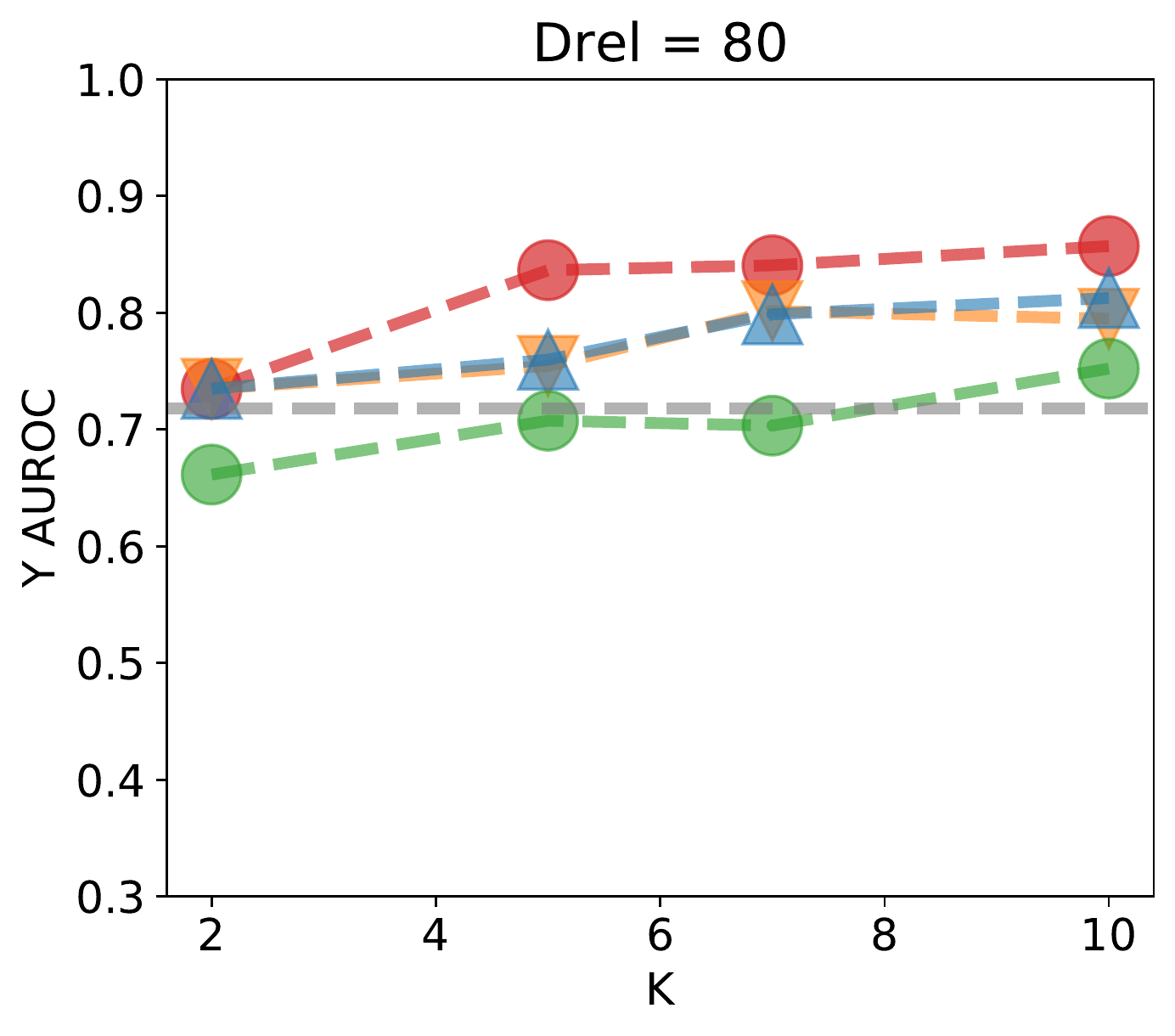}
\includegraphics[width=.3\textwidth]{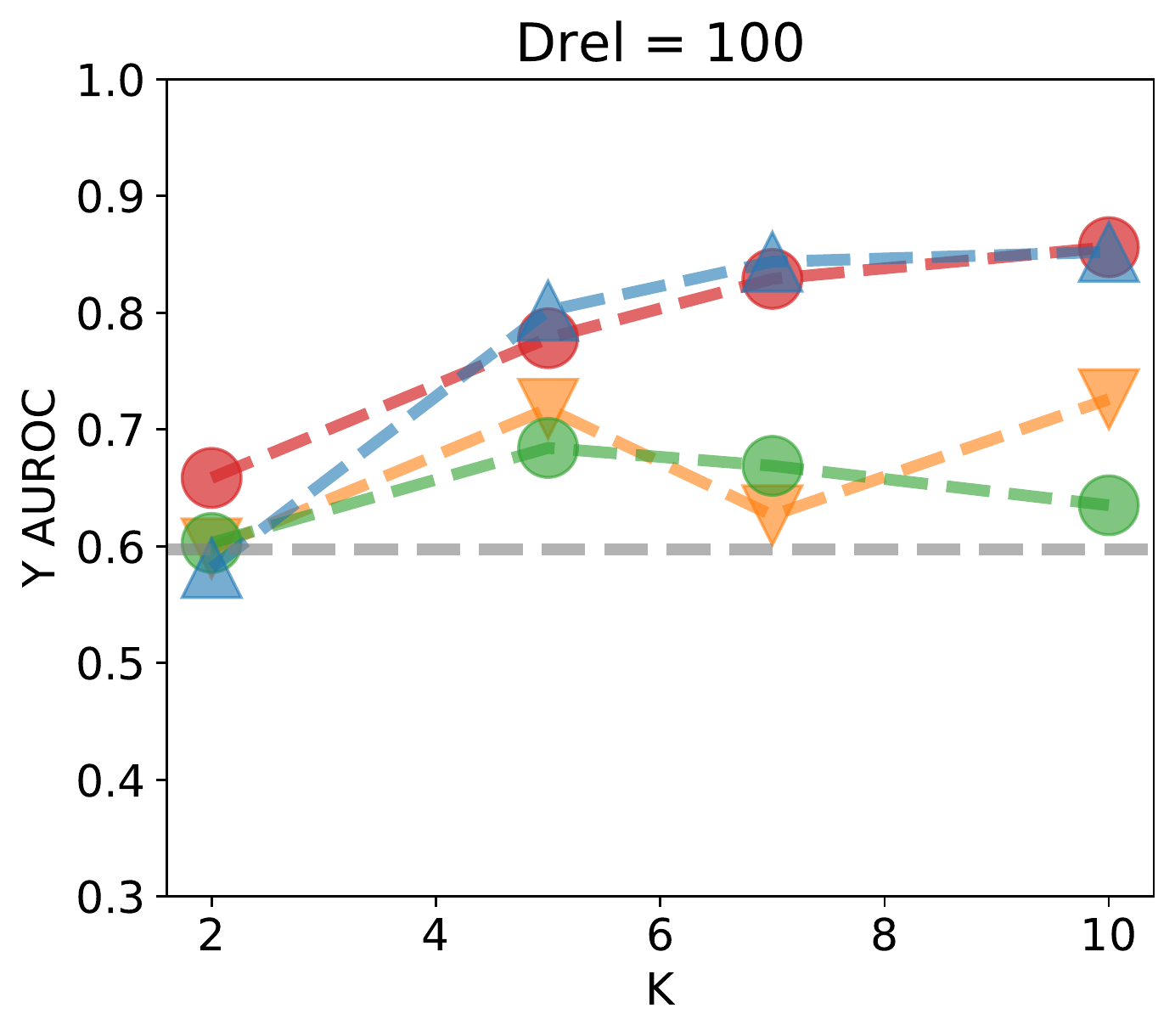}
\includegraphics[width=.3\textwidth]{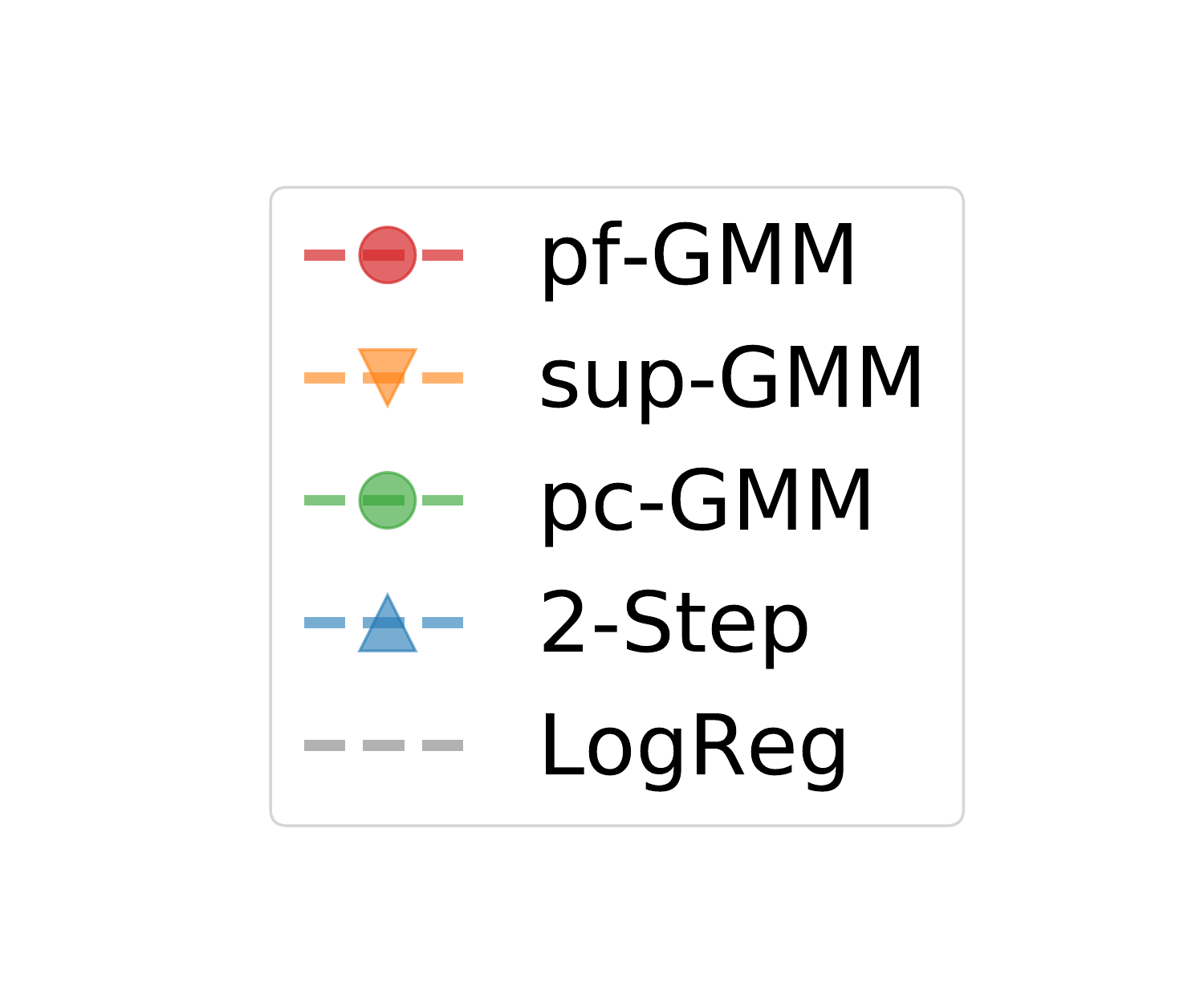}
\caption{\pfGMM\ performance w.r.t. target AUROC on simulated data when we vary the component budget for the mixture models. Each panel corresponds to a dataset with a specified number of relevant dimensions, while keeping the total number of dimensions fixed at 100. Notice our predictive performance is close to optimal (as measured by a discriminative model e.g. Logistic regression).}
\end{figure}

\subsection{\pfHMM\ Simulated Data Experiments}
\begin{figure}[H]
\centering
\includegraphics[width=.3\textwidth]{figures/pfhmm_roc_vs_K_D_20_Drel_2.pdf}
\includegraphics[width=.3\textwidth]{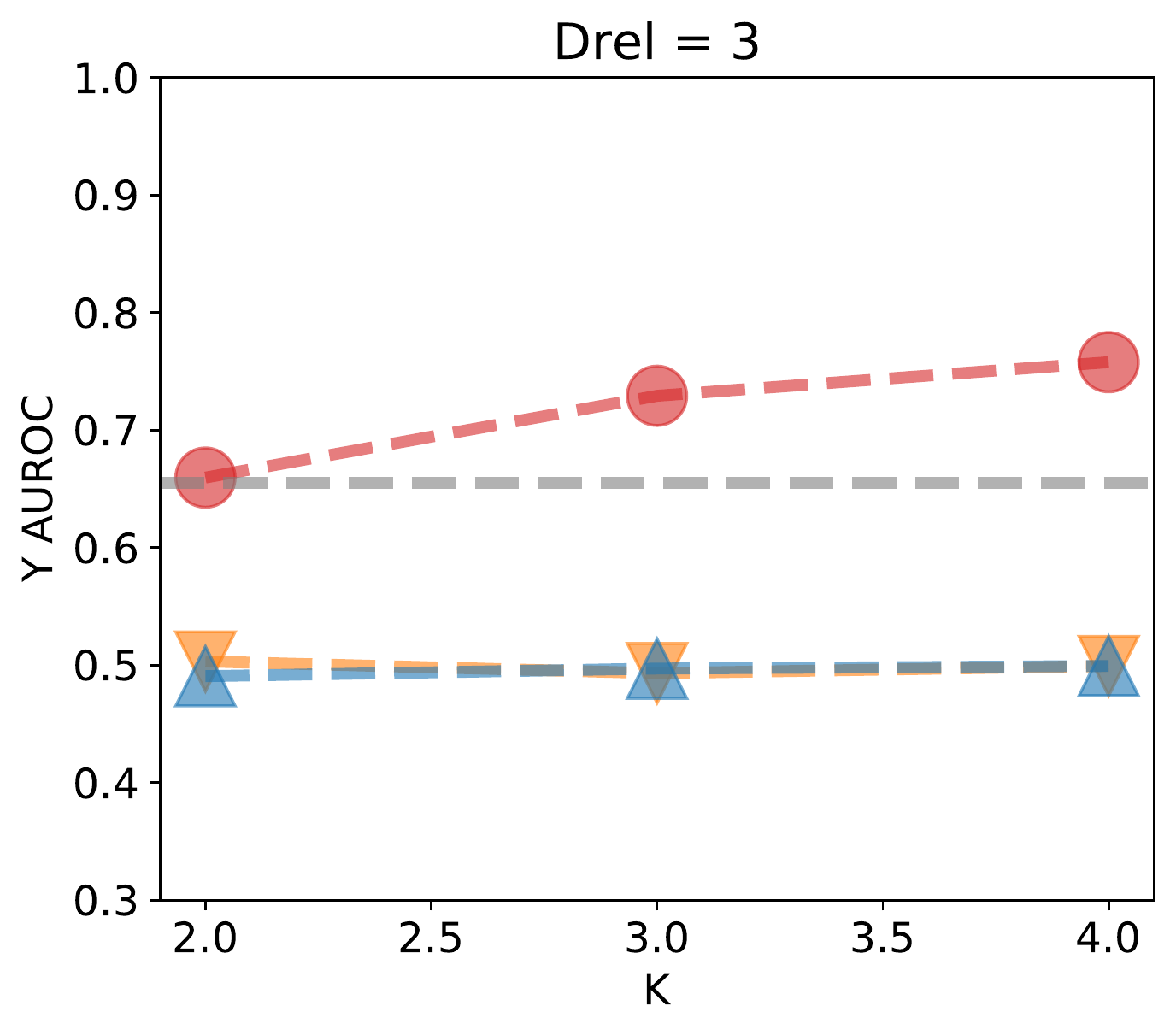}
\includegraphics[width=.3\textwidth]{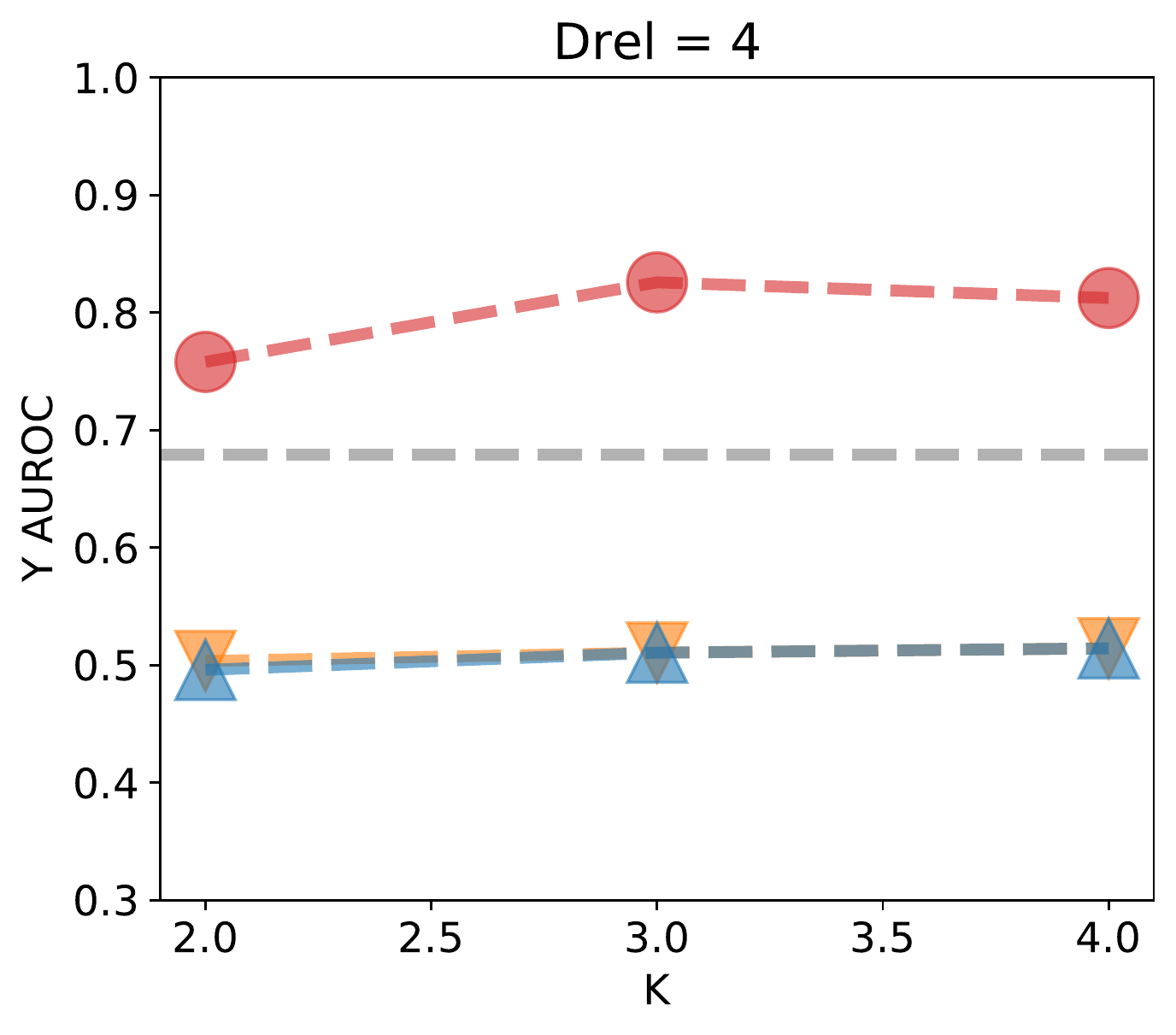}
\includegraphics[width=.3\textwidth]{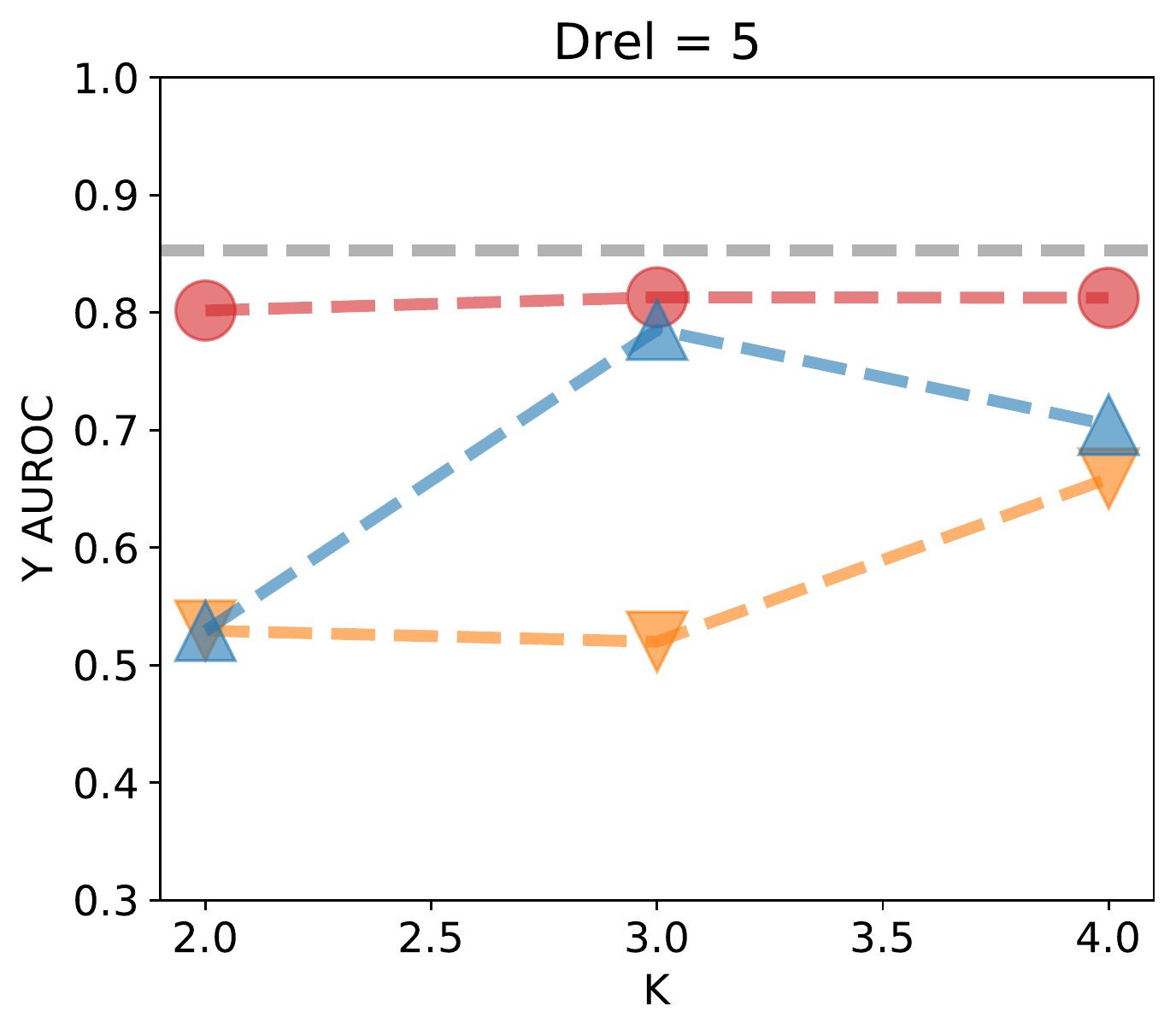}
\includegraphics[width=.3\textwidth]{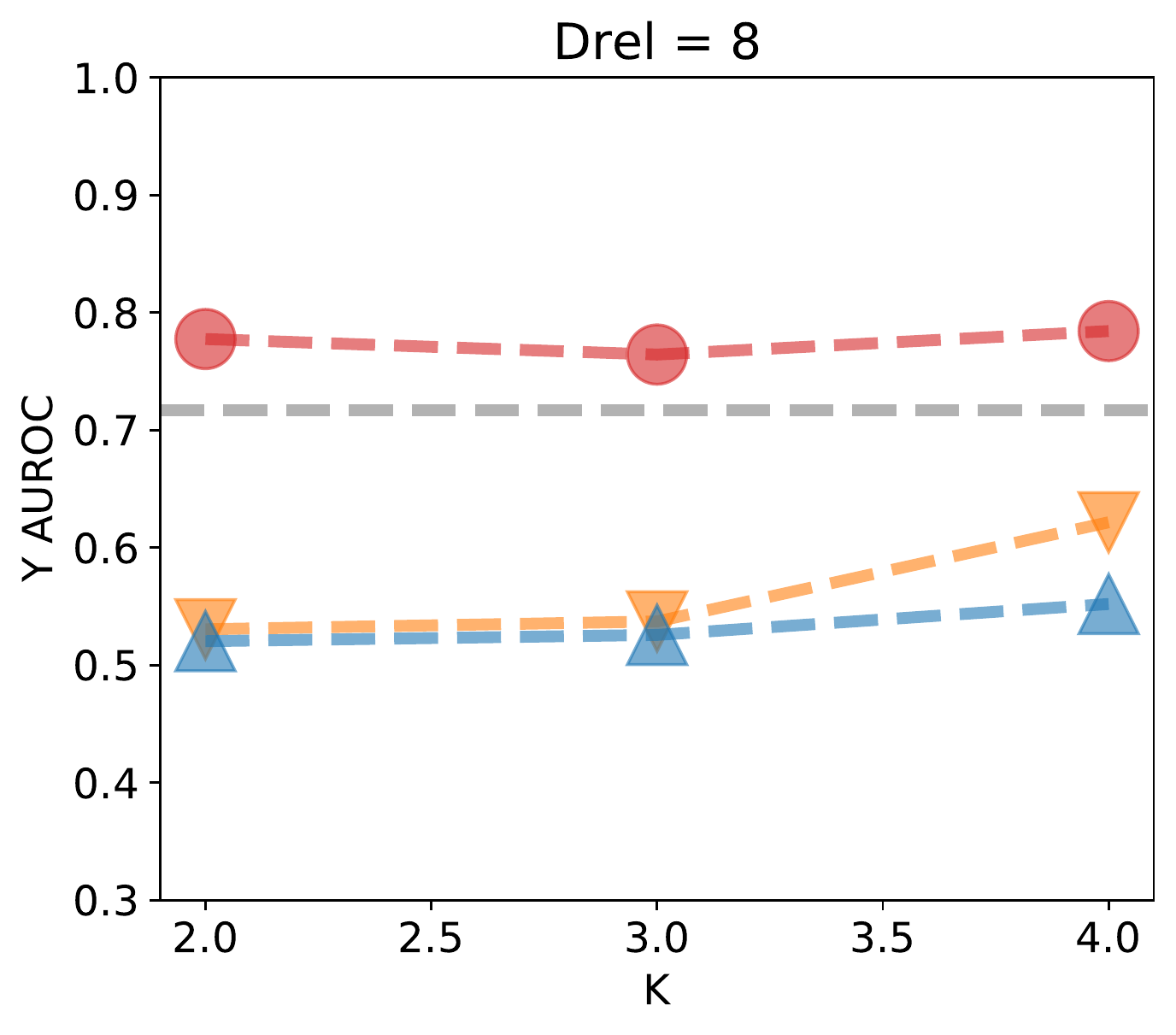}
\includegraphics[width=.3\textwidth]{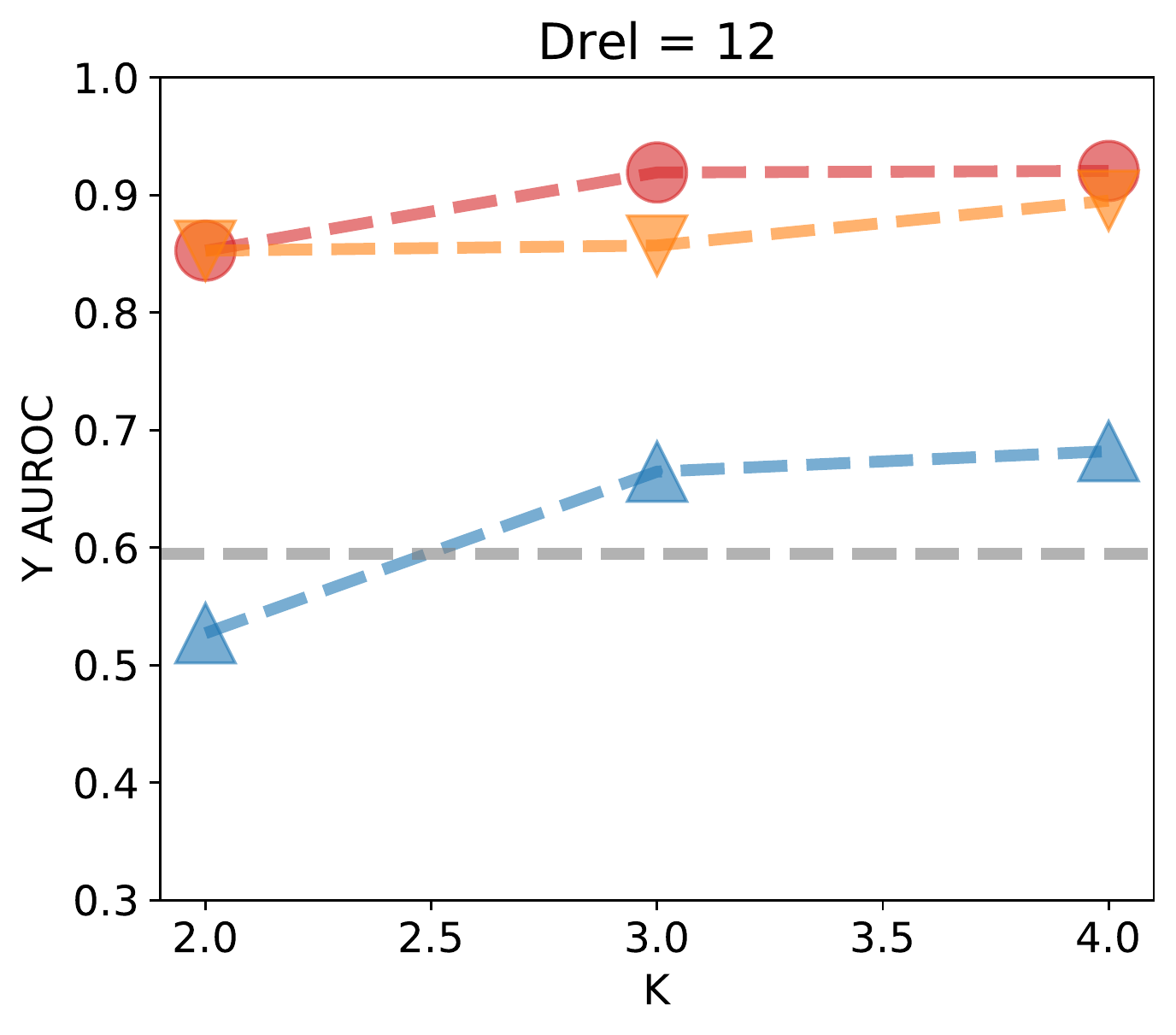}
\includegraphics[width=.3\textwidth]{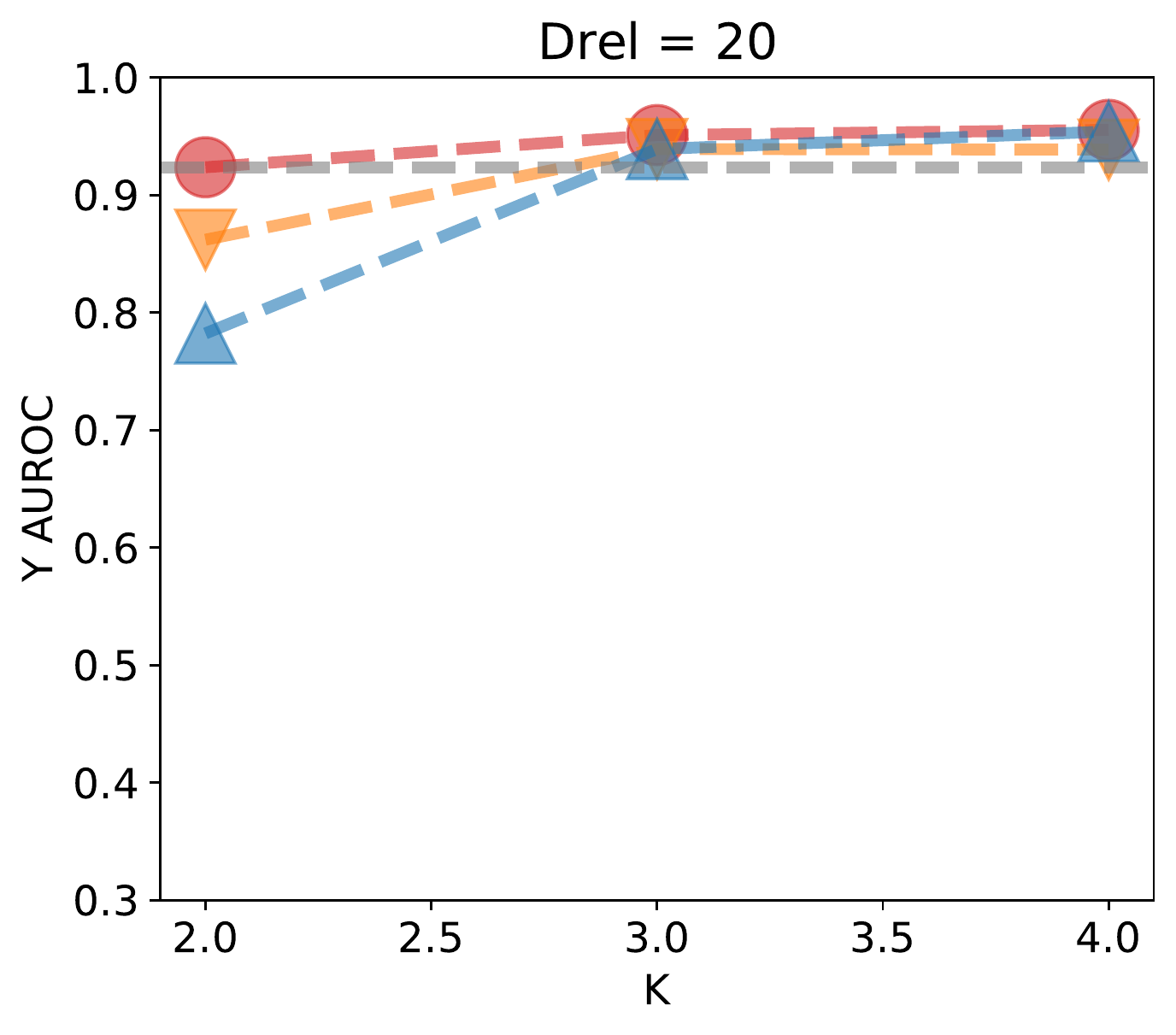}
\includegraphics[width=.3\textwidth]{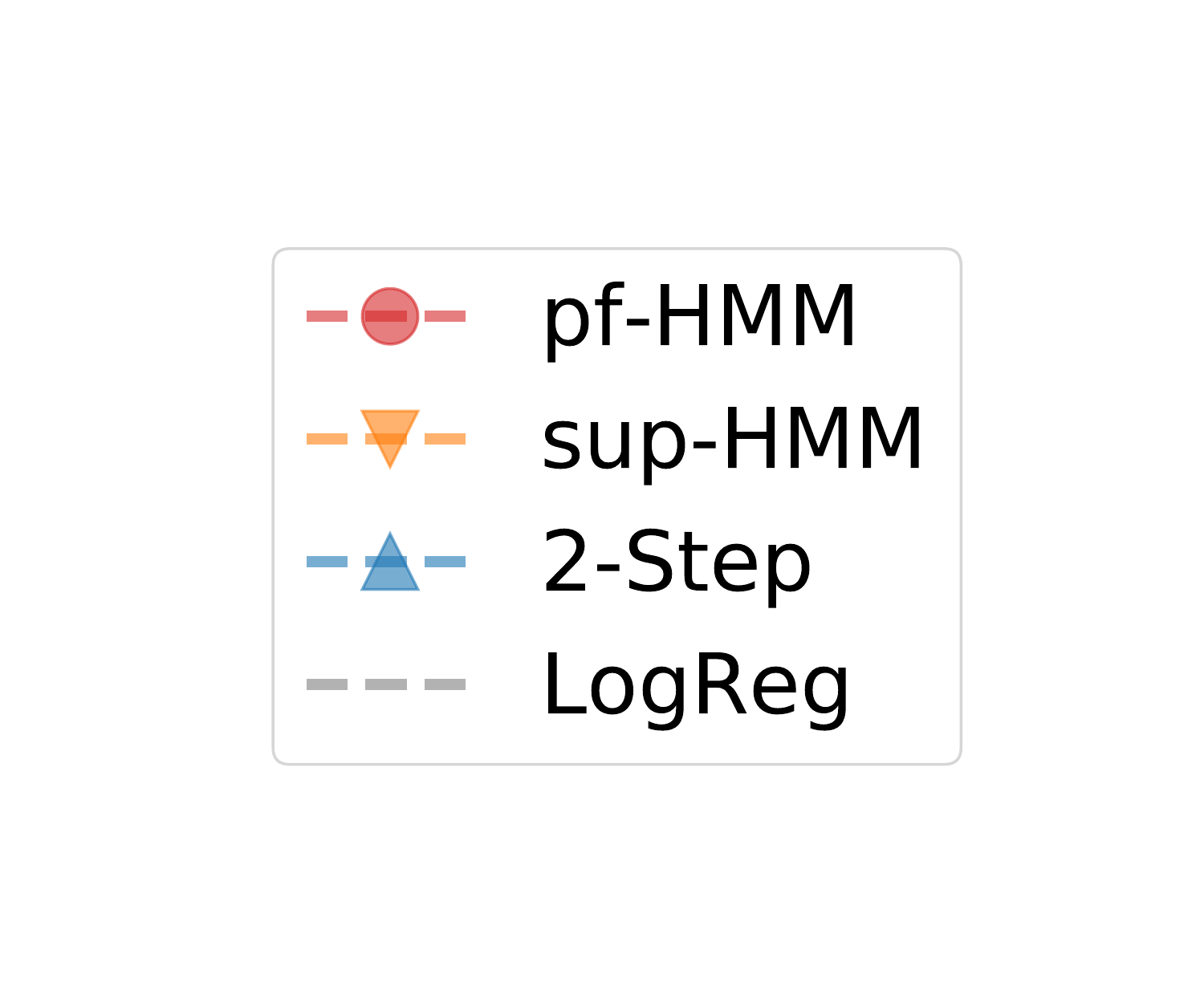}
\caption{\pfHMM\ performance w.r.t. target AUROC on simulated data when we vary the component budget for the mixture models. Each panel corresponds to a dataset with a specified number of relevant dimensions, while keeping the total number of dimensions fixed at 20.}
\end{figure}

\newpage
\section{Experimental Details}

\newcommand{\rel}{\text{Rel}}
\newcommand{\irrel}{\text{Irrel}}

\subsection{Simulated datasets for the \GMM\ experiments}

We follow the following generative process to generate a collection of datasets:
\begin{gather*}
    p_k \sim \textbf{Unif}\parens{\set{0.05, 0.95}} \forall k\\
    [\xseq_1, y] \sim \sum_{k=0}^{K-1} \theta_k^\rel [\Normal{K \cdot \textbf{6}, I}, \Bern{p_k}] \\
    \xseq_2 \sim \sum_{k=0}^{K-1} \theta_k^\irrel \Normal{K \cdot \textbf{6}, I}\\
    \xseq = [\xseq_1; \xseq_2]; \quad \xseq_1 \in \R^{D_\rel}; \quad \xseq_2 \in \R^{D_\irrel} \\
\end{gather*}
where $\textbf{Unif}\parens{\set{a, b}}$ selects either $a$ or $b$ uniformly at random, $\Bern{a}$ is a Bernoulli distribution with parameter $a$, and \textbf{6} is a constant vector. This process ensures that the target $y$ is correlated with the first $D_\rel$ dimensions of the input $\xseq$ but not the last $D_\irrel$ dimensions. The gap between successive components is fixed to 6 so that the clusters are well separated and so that the relevant and irrelevant dimensions have equivalent distributions a-priori.

In our experiments, we set:
\begin{gather*}
    K=10 \\
    D_\rel \in \set{10, 20, 30, 40, 50, 80, 100} \\
    D_\rel + D_\irrel = 100 \\
    \theta^\rel = normalize(0.5 + [0, \dots, K-1]) \\
    \theta^\irrel = normalize(1 + [0, \dots, K-1])
\end{gather*}
where $normalize(\textbf{x}) = \textbf{x} / \norm{\textbf{x}}_1$ makes sure the vectors are valid probability distributions.

This gives us a knob to tweak: $D_\rel$ (number of relevant dimensions). It is common for the input to only have a few relevant dimensions. Therefore, we vary \drel in \set{10, 20, 30, 40, 50, 80, 100}.

\subsection{Simulated datasets for the \HMM\ experiments}

\begin{gather*}
    % p_k \sim  \forall k\\
    [\xseq_1, y] \sim \sum_{k=0}^{K-1} \theta_k^\rel [\Normal{K \cdot \textbf{6}, I}, \Bern{p_k}] \\
    \xseq_2 \sim \sum_{k=0}^{K-1} \theta_k^\irrel \Normal{K \cdot \textbf{6}K, I}\\
    \xseq = [\xseq_1; \xseq_2]; \quad \xseq_1 \in \R^{D_\rel}; \quad \xseq_2 \in \R^{D_\irrel} \\
\end{gather*}

In our experiments, we set:

\begin{gather*}
    K = 4 \\
    p = [0.05, 0.95, 0.05, 0.95] \\
    D_\rel \in \set{2, 3, 4, 5, 8, 12, 20} \\
    D_\rel + D_\irrel = 20 \\
    \theta^\rel = \theta^\irrel = [.1, .2, .3, .4] \\
    A^\rel \gets normalize_{row}( .1 + 
    \begin{bmatrix}
    - z_1 - \\ \dots \\ - z_K -
    \end{bmatrix}
     + \textbf{I}_K)\\
    A^\irrel \gets normalize_{row}( .01 + 
    \begin{bmatrix}
    - z'_1 - \\ \dots \\ - z'_K -    
    \end{bmatrix}
     + \textbf{I}_K)\\
    z_i, z'_i \sim \Cat{\textbf{1}/K}, z_i, z'_i \in \R^K
\end{gather*}
where $normalize_{row}(M)$ applies $normalize$ to each row of matrix $M$. Also, the numbers .1 and .01 are arbitrary choices to differentiate $A^\rel$ and $A^\irrel$, and the simulation would work with different values too.

\subsection{HIV Experiments}
Therapy  for HIV involves administering cocktails of antiretrovirals from  five  classes  namely,  Non-nucleoside Reverse Transcriptase Inhibitors (nnRTIs),  Nucleoside Reverse Transcriptase Inhibitors (nRTIs),  Protease Inhibitors (PIs),  Fusion Inhibitors (FIs), and  Integrase Inhibitors (IIs)  to  bring  the  viral  load  below  detection  limits  ($\leq 40$  copies/ml). We  study 53\, 236 patients with HIV from the EuResist Integrated Database.  Each person has a time-series of average length 16 steps where a time step is approximately 4 months between consecutive treatments. Our task is to predict whether a treatment will bring the viral load below detection limits in the next time-step. Each input contains 138 features including CD4+counts, genetic mutations, treatments in terms of drug classes and lab results. Though it is common to have many genetic mutations, only a few of these may be relevant for inducing drug resistance thus increasing the viral load.

% \vfill

% \end{document}

\end{document}